\documentclass[sn-nature]{sn-jnl}

\usepackage{graphicx}
\usepackage{multirow}
\usepackage{amsmath,amssymb,amsfonts}
\usepackage{amsthm}
\usepackage{mathrsfs}
\usepackage[title]{appendix}
\usepackage{xcolor}
\usepackage{textcomp}
\usepackage{manyfoot}
\usepackage{booktabs}
\usepackage{makecell}
\usepackage{array}
\usepackage{amsmath}
\usepackage{multirow}
\usepackage{algorithm}
\usepackage{algorithmicx}
\usepackage{algpseudocode}
\usepackage{adjustbox}
\usepackage{tabularx}
\usepackage{listings}

\usepackage{xcolor}
\usepackage{geometry}
\newtheorem{theorem}{Theorem}
\newtheorem{proposition}[theorem]{Proposition}%
\newtheorem{definition}{Definition}%

\begin{document}
\title[Article Title]{Key-Embedded Privacy for Decentralized AI in Biomedical Omics}


\author[1,2,3]{\fnm{Rongyu} \sur{Zhang}}\equalcont{Equal contribution.}

\author[4,5,6]{\fnm{Hongyu} \sur{Dong}}\equalcont{Equal contribution.}

\author[3]{\fnm{Gaole} \sur{Dai}}\equalcont{Equal contribution.}

\author[7]{\fnm{Ziqi} \sur{Qiao}}\equalcont{Equal contribution.}

\author[1]{\fnm{Shenli} \sur{Zheng}}

\author[3]{\fnm{Yuan} \sur{Zhang}}

\author[3]{\fnm{Aosong} \sur{Cheng}}

\author[3]{\fnm{Xiaowei} \sur{Chi}}

\author[7]{\fnm{Jincai} \sur{Luo}}

\author[8]{\fnm{Pin} \sur{Li}}

\author[1]{\fnm{Li} \sur{Du}}

\author[11]{\fnm{Dan} \sur{Wang}}

\author*[1]{\fnm{Yuan} \sur{Du}}\email{yuandu@nju.edu.cn}
\equalsup{Equal supervision.}

\author*[9]{\fnm{Xudong} \sur{Xing}}\email{xingxd@big.ac.cn}
\equalsup{Equal supervision.}

\author*[10]{\fnm{Jianxu} \sur{Chen}}\email{jianxu.chen@isas.de}
\equalsup{Equal supervision.}

\author*[3]{\fnm{Shanghang} \sur{Zhang}}\email{shanghang@pku.edu.cn}
\equalsup{Equal supervision.}

\affil[1]{\orgname{School of Electronic Science and Engineering}, \orgname{Nanjing University}, \orgaddress{\city{Nanjing, Jiangsu}, \postcode{210023}, \country{China}}}

\affil[2]{\orgname{Department of Computing}, \orgname{The Hong Kong Polytechnic University}, \orgaddress{\city{Hong Kong}, \postcode{999077}, \country{China}}}

\affil[3]{\orgdiv{State Key Laboratory of Multimedia Information Processing}, \\ \orgname{Peking University}, \orgaddress{\city{Beijing}, \postcode{100871}, \country{China}}}

\affil[4]{\orgdiv{College of Computer Science and Technology}, \orgname{Zhejiang University}, \orgaddress{\city{Hangzhou, Zhejiang}, \postcode{310058}, \country{China}}}

\affil[5]{\orgdiv{College of Engineering}, \orgname{Westlake University}, \orgaddress{\city{Hangzhou, Zhejiang}, \postcode{310058}, \country{China}}}

\affil[6]{\orgname{Zhongguancun Academy}, \orgaddress{\city{Beijing}, \postcode{100094}, \country{China}}}

\affil[7]{\orgdiv{College of Future Technology}, \orgname{Peking University}, \orgaddress{\city{Beijing}, \postcode{100871}, \country{China}}}

\affil[8]{\orgname{China Pharmaceutical University}, \orgaddress{\city{Nanjing, Jiangsu}, \postcode{211122}, \country{China}}}

\affil[9]{\orgdiv{Beijing Institute of Genomics}, \orgname{Chinese Academy of Sciences and China National Center for Bioinformation}, \orgaddress{\city{Beijing}, \postcode{100101}, \country{China}}}

\affil[10]{\orgname{Leibniz-Institut für Analytische Wissenschaften – ISAS – e.V.}, \orgaddress{\street{Bunsen-Kirchhoff-Str. 11}, \city{Dortmund}, \postcode{44139}, \country{Germany}}}

\affil[11]{\orgdiv{Academy of Interdisciplinary Studies}, \orgname{Hong Kong University of Science and Technology}, \orgaddress{\city{Hong Kong}, \postcode{999077}, \country{China}}}

\abstract{The rapid adoption of data-driven methods in biomedicine has intensified concerns over privacy, governance, and regulation, limiting raw data sharing and hindering the assembly of representative cohorts for clinically relevant AI. This landscape necessitates practical, efficient privacy solutions, as cryptographic defenses often impose heavy overhead and differential privacy can degrade performance, leading to sub-optimal outcomes in real-world settings. Here, we present a lightweight federated learning method, INFL, based on Implicit Neural Representations that addresses these challenges. Our approach integrates plug-and-play, coordinate-conditioned modules into client models, embeds a secret key directly into the architecture, and supports seamless aggregation across heterogeneous sites. Across diverse biomedical omics tasks, including cohort-scale classification in bulk proteomics, regression for perturbation prediction in single-cell transcriptomics, and clustering in spatial transcriptomics and multi-omics with both public and private data, we demonstrate that INFL achieves strong, controllable privacy while maintaining utility, preserving the performance necessary for downstream scientific and clinical applications.}

\maketitle

\section{Introduction}\label{sec:introdution}

In the past decade, we have witnessed a tandem fast-growth in bioinformatic algorithms and experimental analytical approaches, which have profound transformation on how we conduct biomedical research towards many diseases. Artificial intelligence (AI) techniques have become an important tool in bioinformatics for many data-driven studies, such as genomic analysis~\cite{przybyla2022new}, spatical proteomics~\cite{mund2022deep}, drug discovery~\cite{askr2023deep}, etc. However, the heterogeneity in data is still one of the primary challenges for many tasks, where large institution-spanning or multi-center studies have become inevitable. The tremendous efforts in data standardization across various scales, from DNA/RNA to proteome and metabolome~\cite{oberacher2020european,perez2025pride}, in the past few years, have paved the way for interoperability, large-scale data integration and consolidation. But, in real-life scenarios, especially in clinical settings, there are still many practical hurdles, such as data privacy, data ownership, governance, human-subjects ethics, intellectual property, etc, combined with stringent and heterogeneous cross-jurisdictional regulations~\cite{voigt2017eu}, which impedes the physically aggregation of representative cohorts and the development of large-scale reproducible, clinically relevant models. This stands in stark contrast to the huge strides in commercial AI models built on non-sensitive data. Consequently, enabling cross-site collaboration without exposing raw data while preserving model utility has become a central prerequisite for moving biomedical AI beyond isolated data silos toward scalable translational impact, motivating the adoption of secure, privacy-preserving distributed learning paradigms.

To alleviate the structural bottlenecks of centralized data sharing and model training, federated learning (FL)~\cite{mcmahan2017communication} was proposed as a privacy-preserving paradigm for cross-institutional learning consolidation without transferring raw data~\cite{dayan2021federated}. In federated learning, participating clients train on sensitive data locally and share only model updates, such as parameters or gradients, with a cloud server to form a global consensus model, thereby preserving data sovereignty and regulatory compliance while integrating diverse, geographically distributed cohorts. This distributed framework has shown considerable promise with translational potential in biomedical research and clinical practice~\cite{cai2025federated}. 

However, federated learning is not always ``leakproof". It is possible that untrusted server may mount data reconstruction attacks and therefore sensitive information could be leaked. Two common techniques to fortify federated learning as a practical and trustworthy solution in biomedicine are cryptographic computation~\cite{joseph2022transitioning} and differential privacy~\cite{islam2022differential}.

\textbf{Cryptographic augmentations}, such as secure multi-party computation (MPC)~\cite{chen2024secure} and multi-party homomorphic encryption (MHE)~\cite{jiang2025towards,froelicher2021truly}, could impose a strong protection against exposing raw data while preserving model accuracy with cryptographic techniques. For example, 1) SDF-ASMC~\cite{chen2024secure} fuses secret sharing, Diffie–Hellman, and functional encryption to provide authenticated, secure MPC without a trusted third party; 2) optimized fully Byzantine-robust homomorphic encryption~\cite{jiang2025towards} could also be used to prevent information leakage while enabling fast, practical aggregation. However, MPC and MHE offer strong privacy with the cost of imposing heavy communication, computation, and storage overheads, scale poorly at large client counts, and centralize risk through a single point of failure and key management, so a breach can compromise the entire system.

\textbf{Differential privacy (DP)}~\cite{dwork2006differential} offers a quantifiable trade-off between utility and privacy by injecting noise during training or release, thereby maintaining protection even if the model is compromised. For example, 1) PrivateKT~\cite{qi2023differentially} transfers knowledge via actively selected small public data under differential privacy, remarkably closing the performance gap to centralized learning; 2) Fed-SMP~\cite{hu2023federated} sparsifies local models before Gaussian perturbation to improve accuracy under the DP condition. At the same time, the nature of injecting noise inevitably comprises the accuracy of the model to a certain extent. Currently, the excellent privacy-preserving performance of DP makes it emerging as a leading complementary safeguard for federated learning, despite the notable degradation in model performance.


From a bioinformatic perspective, most federated learning studies that incorporate the aforementioned privacy-preserving methods remain theoretical or rely on canonical computer vision benchmarks with limited systematic evaluation on noisier and more heterogeneous real-world biomedical data, limiting the practical adoption and generalizability of federated learning in biomedical settings. To our knowledge, only very few privacy-preservation works have examined federated omics applications. For example, FAMHE~\cite{froelicher2021truly} proposed a multi-party homomorphic encryption–based federated analytics system that preserves privacy while accurately reproducing centralized biomedical analyses, including survival analysis and GWAS, across distributed institutions. PPML-Omics~\cite{zhou2024ppml} incorporated decentralized randomization (DR) with DP to investigate cancer classification using TCGA~\cite{tomczak2015review} bulk RNA-seq, clustering in single-cell RNA-seq, and integrative analyses that couple spatial gene expression with tumor morphology from spatial transcriptomics. However, these works all suffer from significant computational and communication overhead or inevitable performance degradation. In conclusion, the scarcity of such domain-grounded development and evaluations further calls for a generalized privacy-preserving federated learning method that can accommodate diverse omics modalities and biomedical application scenarios.

\begin{figure}[htbp]
    \centering
    \includegraphics[width=\textwidth]{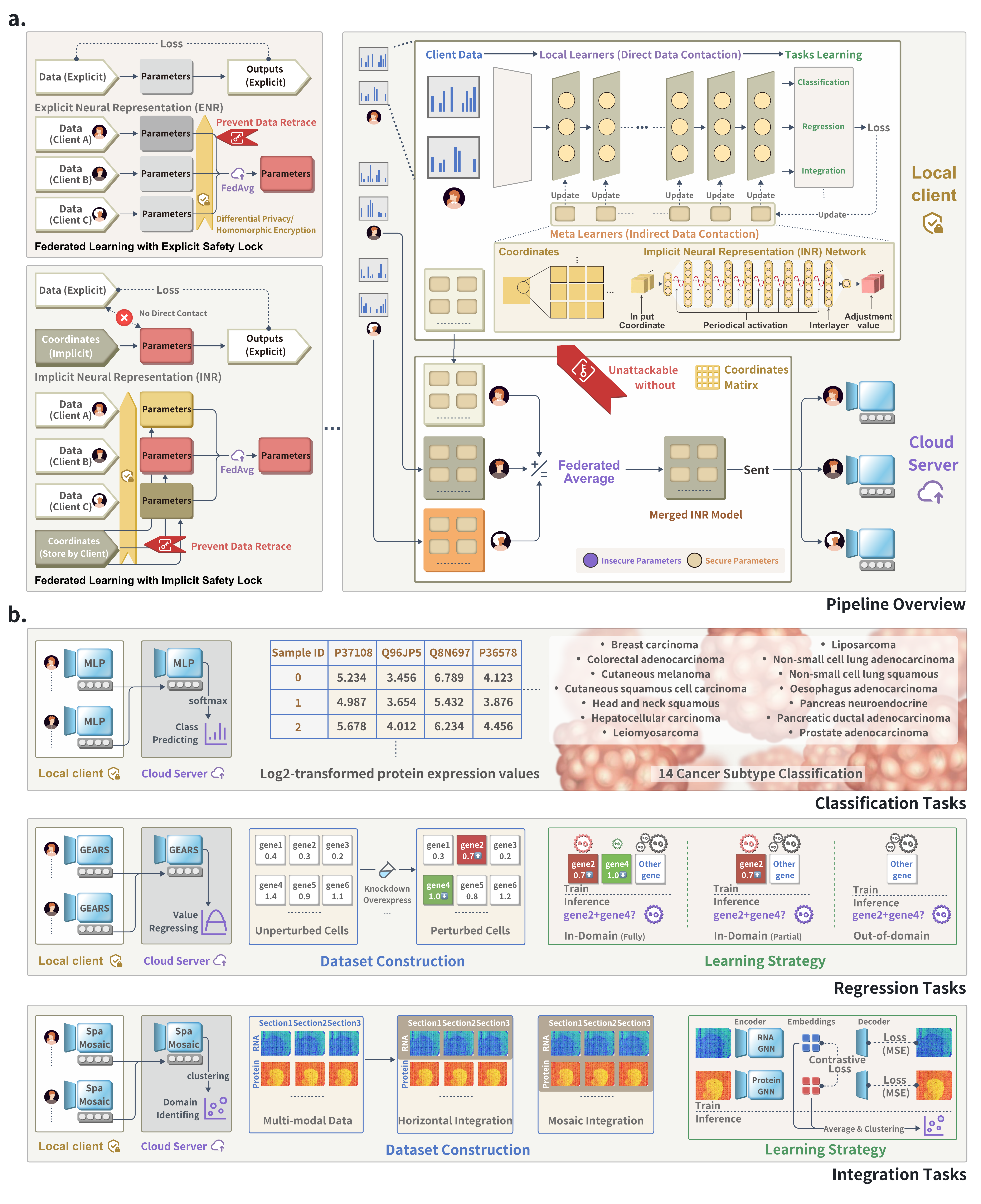}
    \caption{\textbf{Overview of Implicit Neural Federal Learning (INFL).}
\textbf{a} The left column provides a concise comparison between traditional Federated Learning using explicit neural networks and the proposed approach based on implicit neural networks. In explicit settings, user data utilized for training local models may be vulnerable to attacks if not protected by privacy-preserving mechanisms such as differential privacy (DP) or homomorphic encryption (HE). In contrast, in the implicit framework, the model is inherently secure because it does not directly access raw user data. The right column illustrates the detailed workflow of INFL: each user train Meta Learners alongside the Local Learner and uploads only the Meta Learner parameters. These are then aggregated at the server via the Federated Averaging algorithm.
\textbf{b} The top panel demonstrates the application of INFL to cancer subtyping based on protein expression levels. A multilayer perceptron (MLP) classifier (n=14) is initialized on each client and serves as the Local Learner. The middle panel adopts the GEARS model as the backbone architecture to perform gene expression level regression under various perturbations, evaluating both in-domain and out-of-domain gene predictions. The bottom panel employs SpaMosaic as the Local Learner for multi-modal spatial transcriptomics integration, supporting both horizontal integration (same modality across different sections) and mosaic integration (different modalities across sections).
}
    \label{fig:figure1}
\end{figure}

To address the aforementioned challenges, we propose a lightweight and privacy-preserving Implicit Neural Federated Learning (INFL) approach, which incorporates a plug-and-play encryption scheme built upon Implicit Neural Representations (INRs)~\cite{sitzmann2020implicit,dai2025implicit} integrated into various linear layers within the clients' local models, as shown in \textbf{Figure~\ref{fig:figure1}-a}. These modules function analogously to low-rank adaptation (LoRA)~\cite{hu2022lora}, co-training, and aggregating with the backbone while embedding a secret key into the model architecture. However, unlike LoRA that relies on high-level features as input, INFL introduces self-defined coordinate keys as inputs to the INR. These keys act as private identifiers, enabling the model to learn its task while conditioning on these unique coordinates during training. The global model parameters serve as a shared "global key," while the private coordinate keys act as "local keys." During inference, an authorized user with both the global parameters and the correct private key can generate accurate predictions. In contrast, an unauthorized attacker with only the global parameters would be forced to guess the key, resulting in scrambled inputs to the INR. This mismatch disrupts the INR modulation process, leading to corrupted and non-functional model outputs.

We evaluate INFL on a diverse set of representative omics analyses tasks: cohort-scale classification in bulk proteomics, regression for perturbation prediction in single-cell transcriptomics, and clustering in spatial transcriptomics and multi-omics. As shown in \textbf{Figure~\ref{fig:figure1}-b}, representative deep learning models are tested accordingly. We demonstrate that INFL outperforms the state-of-the-art federated learning methods, underscoring its versatility in jointly preserving model performance without sacrificing computation and communication overheads, and providing strong privacy protection compared to traditional federated learning methods.

\section{Results}\label{sec:results}

The overall framework of INFL is illustrated in \textbf{Figure~\ref{fig:figure1}}. Assume there are $K$ different collaborating clients ($C_1, C_2, ..., C_K$), e.g., different participating hospitals, each with their own private data ($D_1, D_2, ..., D_K$) conforming to the same data format and each with their own private computing resource, i.e., their own model acting as local learners ($LocL_1, LocL_2, ..., LocL_K$). It is important to note that INFL is agnostic to the underlying deep learning models (e.g, a Transformer model for regression, a ResNet for classification, etc.), but they models have to be consistent (i.e., using the same architectural design) across all local learners. As soon as one client, say $C_i$, has prepared its private data $D_i$, the local learner $LocL_i$ can start training.



Subsequently, multiple INR modules are attached to different layers of each local learner $LocL_i$ to function as meta learners $MetL_i$. When training $LocL_i$, client data $D_i$ are directly fed into $LocL_i$, completing a standard training pipeline. Simultaneously, the meta learner $MetL_i$ receives either a self-defined key or an authentically generated one as input. This design prevents direct exposure of client data to the meta learners while preserving their capacity to contribute to the learning process. This key-driven design enables meta learners to process encrypted representations rather than raw client data, ensuring strong privacy protection. At the same time, the meta learners contribute to improving the global model by providing additional structured information, enhancing overall performance during federated aggregation.

After local training on local clients $C_i$, model weights of $LocL_i$ are transmitted to the cloud server. The weights from all (or a subset of all) local learners are aggregated using the FedAvg~\cite{mcmahan2017communication} algorithm to update the parameters of the global model on the cloud, which are then sent back to each client $C_i$ to update the weights of $LocL_i$ and perform further local training with $D_i$. Since the meta learners are designed to learn indirectly from client data, neither additional computational overhead (comparing to employing homomorphic encryption) nor performance degradation (comparing to differential privacy) is observed (see Supplementary Figure-1 for sanity checks.)


Formal proof of our INR-based framework as a cryptographic lock is provided in Section~\ref{sec:theoretical_analysis}. Here, we emphasize that INFL is a general model-agnostic mechanism, compatible with standard architectures and federated pipelines, yielding strong performance across heterogeneous biological settings. Concretely, we demonstrate the effectiveness using distinct base models under varied federated configurations: in Section~\ref{sec:results-Cancer Subtyping}, a multi-layer perceptron (MLP) model for proteomics-based cancer subtyping; in Section~\ref{sec:results-Gene Perturbation}, GEARS model for transcriptomics perturbation response; and in Sections~\ref{sec:results-Multi-Omics Integration} and~\ref{sec:results-Multi-Omics Integration-Mosaic}, SpaMosaic model for spatial transcriptomics and spatial multi-omics integration, respectively. Notably, the SpaMosaic application in Section~\ref{sec:results-Multi-Omics Integration-Mosaic} operates on inherently non-iid, multi-modal data (ADT and RNA), illustrating that INFL accommodates biological heterogeneity without bespoke, task-specific tuning across both spatial and non-spatial omics.

In the Supplementary (Section~\ref{sec:sup}), we further stress-test federated settings on standard vision benchmarks with varying client number, participation ratio, INR size, and non-iid severity, demonstrating robustness to data heterogeneity. These controlled ablations complement our biological results and support INFL as a unified privacy-preserving layer that generalizes to non-iid, multi-center–like regimes, exemplified by the ADT–RNA integration in Section~\ref{sec:results-Multi-Omics Integration-Mosaic}.


\begin{figure}[htbp]
    \centering
    \includegraphics[width=\textwidth]{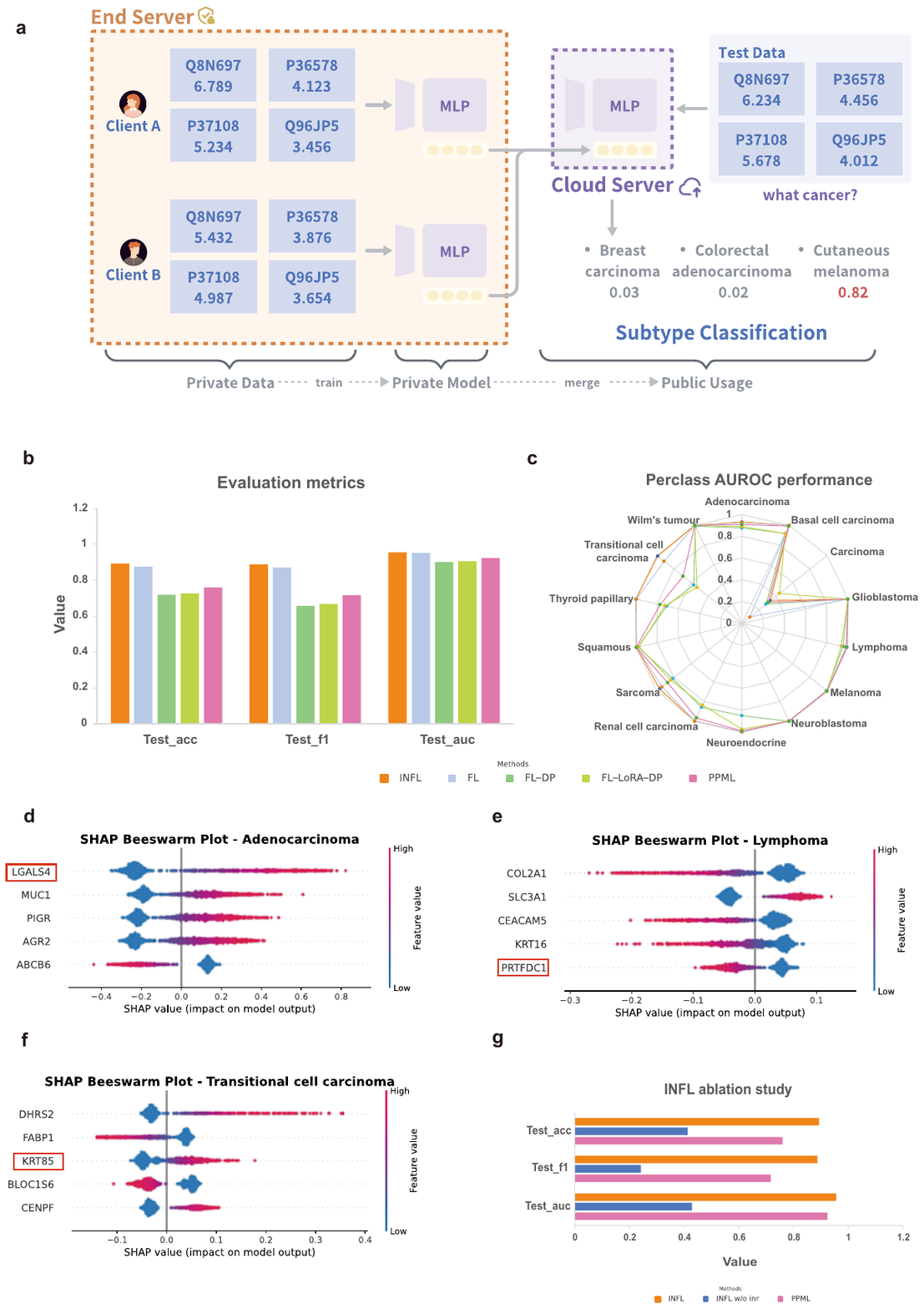}
    \caption{\textbf{Experimental results on bulk proteomics cancer subtyping task.}
\textbf{a} Experimental design of an algorithm for the cancer subtyping task, including training data partitioning, the training workflow, the final inference process, and downstream tasks.
\textbf{b} Overall classification performance evaluated by quantitative metrics. Accuracy, F1-score, and AUROC are reported, with each metric representing the macro-average across all 14 class labels.
\textbf{c} Radar chart displaying the individual AUROC metrics for each of the 14 class labels.
\textbf{d-f} Interpretability analysis of the classification model using Shapley values. For each subtype, the top 5 most relevant proteins are shown. The horizontal axis represents the magnitude of the Shapley value, while the color intensity in each beeswarm plot indicates the expression level of the corresponding protein. A positive Shapley value for a highly expressed protein suggests a positive correlation with the cancer subtype, whereas a negative value indicates a negative correlation.
\textbf{d} Interpretability analysis for Adenocarcinoma.
\textbf{e} Interpretability analysis for Lymphoma.
\textbf{f} Interpretability analysis for Transitional cell carcinoma.
\textbf{g} Ablation study of INFL. Experiments were conducted after removing the INR module from the original model during the inference, and the overall classification performance was evaluated.
}
    \label{fig:figure2}
\end{figure}

\subsection{INFL Excels in Cancer Subtyping for Bulk Proteomics}\label{sec:results-Cancer Subtyping}


We start testing our INFL framework with a simple, yet widely applicable task: cancer subtyping using bulk proteomics data from a large population cohort. Remarkable results have been achieved in cancer subtyping by measuring the human proteome, significantly improving diagnostic and therapeutic accuracy~\cite{sun2025protein} in clinical practices. With the accumulation of large-cohort population data, it has become possible to integrate these data to train a universal model for cancer diagnosis and treatment, which is of immense clinical application value. However, much of this cohort data comes from different hospitals. Due to concerns about patient privacy and the sensitivity of biological data, hospitals face significant challenges in sharing data. Federated learning offers a viable solution to this problem. By training cancer subtyping models locally, followed by the transfer, mixing, and iterative training of model parameters, the final model can achieve accurate and generalizable subtyping without the data ever leaving the respective hospitals, thus fulfilling the ultimate objective.

We adopted the data released from~\cite{cai2025federated}, containing a large-cohort bulk proteomics data from a large population cohort, comprising a total of 1207 samples. Each sample contained measurements for 9101 proteins and was associated with a corresponding cancer type label (15 classes in total). As shown in \textbf{Figure~\ref{fig:figure2}a}, we assumed the presence of five clients and partitioned the dataset into ten equal shards and assigned one shard to each of ten clients. We employed an MLP as the core classification model. Various federated learning strategies were applied to this base model to initiate federated learning training. During each communication round, we sampled five clients uniformly at random to participate in federated training. Unless otherwise noted, the globally aggregated model obtained after the final round was used for inference. During the inference stage, the model takes the protein expression values of a sample as input, predicts the probability of each cancer subtype, and selects the subtype with the highest probability as the final classification result. The predicted labels are then compared with the true labels to calculate performance metrics.

We first used quantitative metrics to evaluate the classification performance of each federated learning method. As shown in \textbf{Figure~\ref{fig:figure2}b}, we reported the overall accuracy, F1-score, and AUROC across all 14 labels. The results indicate that INFL performs slightly better than traditional FedAvg federated learning strategies (i.e., the ProCanFDL method from~\cite{cai2025federated}, where the dataset was originally used for) and significantly outperforms other federated learning strategies (such as FedAvg with differential privacy (FL-DP), FedAvg with low-rank adapter and differential privacy (FL-LoRA-DP), and PPML-Omics (PPML)). We further investigated the AUROC for each individual cancer subtype and plotted the radar chart in \textbf{Figure~\ref{fig:figure2}c}. The results show that INFL achieves the best performance across most cancer types, except for a very few (such as Carcinoma), demonstrating the generalizability of INFL's effectiveness.

Subsequently, we conducted additional application-appropriate validation, i.e., employing analytical methods to validate the biological significance of INFL's classification results. Specifically, we calculated Shapley values~\cite{shapley1953value} based on INFL's classifications to determine the contribution of each feature to the prediction result. We selected the top 5 contributing features for each cancer subtype and generated SHAP beeswarm plots to interpret representative examples. As shown in \textbf{Figure~\ref{fig:figure2}d-f}, the model results indicate a positive correlation between the LGALS4 protein and Adenocarcinoma. As reported in the literature, LGALS4 (Galectin-4) is a lectin protein typically expressed in normal epithelial cells of tissues like the gastrointestinal tract. Its expression is upregulated in various adenocarcinomas, particularly colorectal cancer (CRC). It is involved in cell adhesion, proliferation, and signaling, and its high expression is associated with tumor progression and metastatic potential~\cite{hayashi2013galectin}. Furthermore, the PRTFDC1 protein showed a negative correlation with Lymphoma. As previously studied~\cite{huang2025phosphoribosyl}, PRTFDC1 is a frequent deletion site in lymphoma, the restoration of whose expression in lymphoma cell lines significantly inhibits cell proliferation and induces apoptosis, directly demonstrating its tumor suppressor function in lymphoma. Furthermore, KRT85 also showed a positive correlation with Transitional cell carcinoma. Prior works~\cite{lopez2021molecular} have shown its role as an oncogene in certain cancers. It is highly expressed in the basal/squamous subtype of urothelial carcinoma, which is typically associated with a poorer prognosis. In summary, the interpretability analysis shows that the model not only achieves accurate classification but also precisely identifies proteins closely related to each specific subtype. The observed correlations are consistent with the literature, indicating model's biological relevance.


Additionally, we conducted an ablation study to assess the privacy-preserving capabilities of INFL. Notably, when unauthorized attackers input random coordinates without the correct key, the model collapsed, outputting N/A and failing to generate meaningful results. Under an even more extreme setting where the INR module was removed during inference (Figure~\ref{fig:figure2}g), accuracy dropped far below PPML and other baselines. These results highlight the necessity of INFL’s key-conditioned mechanism for robust privacy while sustaining performance. Importantly, the same mechanism disrupts gradient-based attacks in federated training: with output $\mathbf{y}'=\alpha(\mathbf{W}\mathbf{x}+\mathbf{b})+(1-\alpha)\mathbf{\Delta}(\pi)$ and $\mathbf{\Delta}i(\pi)=\Phi{\theta}(\gamma(C(\pi(i))))$, an adversary who assumes the identity key computes INR gradients at mismatched coordinates, effectively replacing $\partial_{\theta}\Phi_{\theta}(\gamma(C(\pi(i))))$ by $\partial_{\theta}\Phi_{\theta}(\gamma(C(i)))$. For high‑frequency INRs, Jacobians at distinct coordinates are weakly correlated, yielding a non‑vanishing gradient gap $\mathbb{E}!\big[|\nabla_{\theta}L(\pi)-\nabla_{\theta}L(\pi_A)|2^2\big]\gtrsim 2(1-\alpha)^2\sigma{\mathrm{J}}^2\sum_i \mathbb{E}[(\partial \ell/\partial y'i)^2]$ and driving the expected cosine similarity toward zero. Thus, gradient matching and model inversion lose the one‑to‑one alignment needed for reconstruction, and adversarial updates deviate from the true descent direction; an identical permutation mismatch propagates to $\nabla{\mathbf{W}}L$ via backpropagation. Full derivations and extensions are provided in Section~\ref{sec:theoretical_analysis}, which also explains why INFL retains its privacy‑preserving behavior under such extreme circumstances.

\begin{figure}[htbp]
    \centering
    \includegraphics[width=\textwidth]{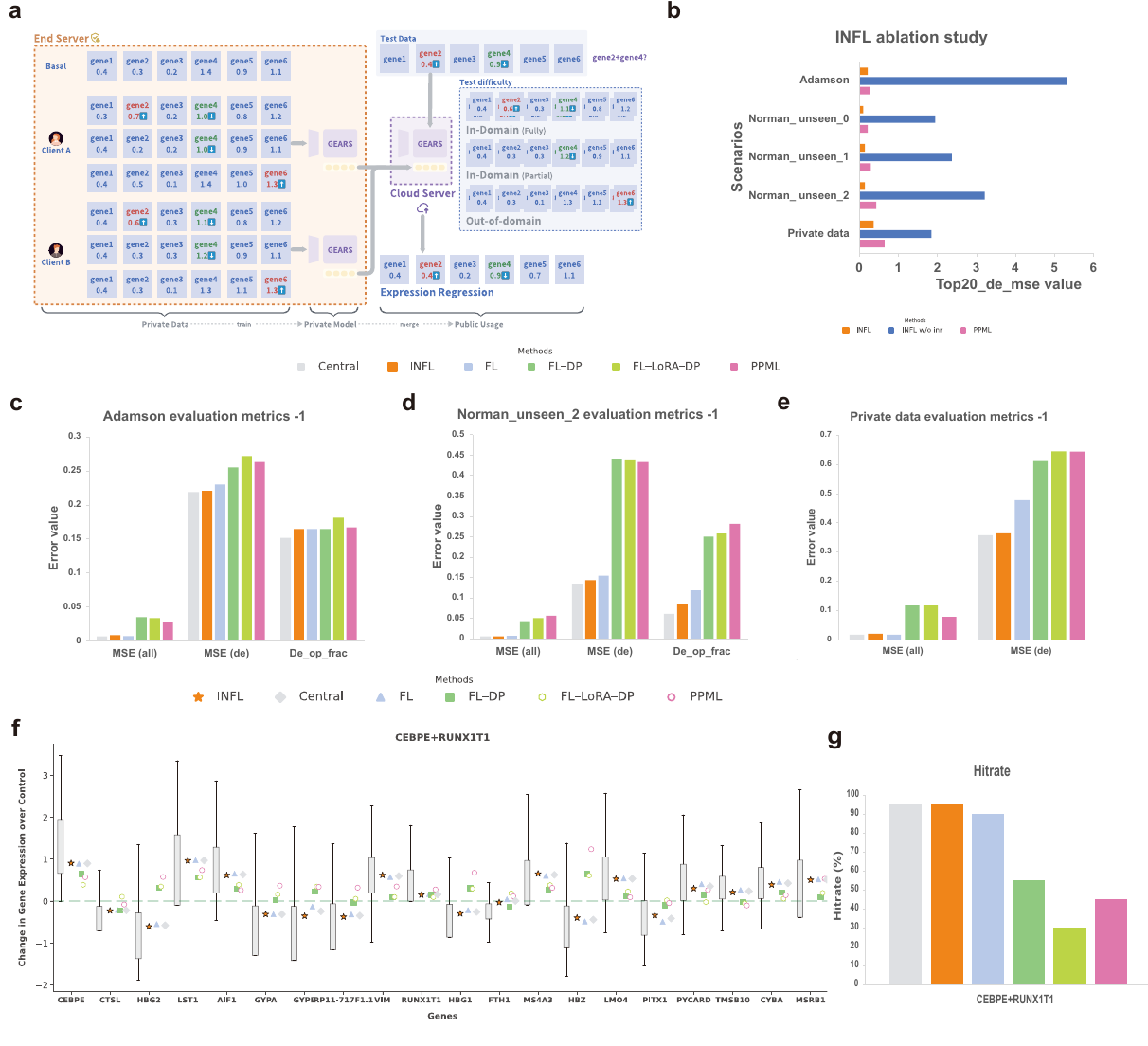}
    \caption{\textbf{Experimental results on the single-cell transcriptomic perturbation prediction task.}
\textbf{a} Experimental design of algorithm for the single-cell transcriptomic perturbation prediction task, including training data partitioning, training workflow, final inference process, and downstream tasks.
\textbf{b} Ablation study of INFL. Experiments were conducted after removing the INR module from the original algorithm, comparing the MSE metric for the top 20 differentially expressed genes across various scenarios.
\textbf{c-e} Model perturbation prediction performance characterized by quantitative metrics. For each scenario, we selected three metrics to demonstrate the performance of each method: the MSE for all genes (MSE(all)), the MSE for the top 20 differentially expressed genes (MSE(de)), and the fraction of top 20 differentially expressed genes with opposite prediction direction (De\_op\_frac). For all three metrics, lower values indicate better performance. \textbf{c} Performance on the Adamson dataset. \textbf{d} Performance on the Norman dataset under the out-of-domain scenario (unseen\_2). \textbf{e} Performance on the private dataset.
\textbf{f} Qualitative case study showing model perturbation performance. The dual-gene perturbation (\textit{CEBPE + RUNX1T1}) from the Norman dataset under the out-of-domain scenario is shown. The green dotted line shows mean unperturbed control gene expression. Boxes indicate experimentally measured differential gene expression after perturbing the gene combination \textit{CEBPE} and \textit{RUNX1T1}. Different symbols shows the mean change in gene expression predicted by different methods when they have not seen \textit{CEBPE} nor \textit{RUNX1T1} experimentally perturbed at the time of training. Whiskers represent the last data point within 1.5× interquartile range.
\textbf{g} Perturbation performance quantitatively displayed using hit rate. A prediction is considered accurate if the mean change in gene expression predicted by a method falls within the boxes range. The hit rate is calculated by summarizing the prediction accuracy across all 20 genes.}
    \label{fig:figure3}
\end{figure}

\subsection{INFL Excels in Perturbation Response Prediction for Single-Cell Transcriptomics.}\label{sec:results-Gene Perturbation} 

We further tested our INFL framework on the task of single-cell transcriptomics perturbation response prediction, where a specialized bioinformatic model is commonly used~\cite{roohani2024predicting}. The single-cell transcriptomics-based prediction of cellular responses to drug perturbation is one of the foundational tasks of the ``Virtual Cell" model~\cite{bunne2024build}, with the promise of in-silico drug screening and great clinical potentials. With the increasing availability of data, it has become possible and highly applicable to aggregate perturbation data from different intervention conditions to train a unified model.

Specifically, the perturbation prediction task refers to training a model on a dataset of triplets consisting of control group gene expression data, perturbation condition, and post-perturbation gene expression data. Subsequently, during testing, the model predicts gene expression under other perturbation conditions, which falls under the regression task in machine learning. For this task, we adopted the commonly used Adamson dataset~\cite{adamson2016multiplexed} and Norman dataset~\cite{norman2019exploring}, together with an additional in-house dataset (see Methods), as a representative diverse data collection to evaluate the model's accuracy and generalizability. The Adamson dataset and our in-house dataset contain only single-gene perturbations, while the Norman dataset contains dual-gene combination perturbations. We assumed the presence of ten clients and partitioned the dataset into ten equal shards, assigning one shard to each of the ten clients. GEARS, a widely recognized perturbation prediction algorithm~\cite{roohani2024predicting}, is adopted as the core model. Various federated learning strategies were applied to this baseline algorithm to initiate federated learning training. During each communication round, we sampled five clients uniformly at random to participate in federated learning. Unless otherwise noted, the globally aggregated model obtained after the final round was used for inference. During the inference stage, the model predicts gene expression under a previously unseen perturbation condition, generating predicted expression values for various genes. These predictions are then compared with the values obtained from real experiments to calculate quantitative metrics.

As shown in \textbf{Figure~\ref{fig:figure3}c-e} and \textbf{Extended Data Figure~\ref{fig:figure3-extend-1}a-b}, we first calculated the mean squared error (MSE) for all genes and the MSE for the top 20 differentially expressed genes. We also calculated the proportion of incorrectly predicted directions for the top 20 differentially expressed genes (where a predicted direction is incorrect if the model predicts downregulation while the true change is upregulation). For all three metrics, lower values are better. It is important to note that for the Adamson and our in-house datasets, which contain only single-gene perturbations, testing was performed on genes not present in the training set. For the Norman dataset, the situation is more complex. We thoroughly considered the concepts of in-domain and out-of-domain generalisation in machine learning~\cite{zhou2022domain}. Taking the prediction of a dual-gene perturbation (\textit{X+Y}) as an example: if single-gene perturbation data for both genes \textit{X} and \textit{Y} exist in the training set, it is considered in-domain (full); if perturbation data for only one gene (e.g., only \textit{X}) exists, it is considered in-domain (partial); if perturbation data for neither \textit{X} nor \textit{Y} exists, it is considered out-of-domain. These are represented in the figures as unseen\_0, unseen\_1, and unseen\_2, respectively. It can be observed that INFL performs on par with the centralised algorithm across all scenarios, slightly outperforms standard federated learning methods, and far surpasses other comparative algorithms (such as FL-DP, FL-LoRA-DP, PPML) in various settings. We also calculated the Pearson correlation coefficient (PCC) for all genes, the PCC for the top 20 differentially expressed genes, and the Pearson\_delta (used to measure the change) to demonstrate the superiority of our method, with conclusions consistent with those from the MSE metrics (\textbf{Extended Data Figure~\ref{fig:figure3-extend-1}c-g}).

In addition to quantitative metrics, we further qualitatively visualized the perturbation responses for certain important cases. For instance, we selected the \textit{CEBPE + RUNX1T1} combined perturbation, which belongs to the out-of-domain (unseen\_2) scenario (\textbf{Figure~\ref{fig:figure3}f-g}). This combined perturbation can relieve myeloid differentiation block, inhibit leukemia stem cell properties, induce apoptosis, and remodel chromatin structure, representing an important strategy for reversing the leukemic phenotype~\cite{derevyanko2025fusion}. It can be observed that compared to other methods, INFL's predictions for the top 20 differentially expressed genes are quite accurate, both in terms of the correct trend and the magnitude of change. We also added a hit rate metric to quantitatively measure prediction accuracy (Methods). If the mean change in gene expression predicted by INFL falls within the boxes range, the prediction is considered accurate. The overall prediction performance across the 20 genes is summarized to calculate the hit rate value. We found that INFL's predictions are on par with the centralised algorithm, reaching 95\%, significantly surpassing other federated learning algorithms. We further demonstrated the method's superiority with more cases (\textit{FARSB+ctrl}, \textit{ZNF318+FOXL2}, \textit{MAP2K6+IKZF3}, \textit{YWHAE+ctrl}) on other datasets and scenarios (\textbf{Extended Data Figure~\ref{fig:figure3-extend-2}a-e}).

Again, we conducted an ablation study to assess the privacy-preserving capabilities of the INR module (\textbf{Figure~\ref{fig:figure3}b}). When the INR module was removed during the inference, the MSE for the top 20 differentially expressed genes across various scenarios increased dramatically compared to the original model and performed significantly worse than PPML. This demonstrates that without the INR module or correct key, the model fails to maintain its performance, highlighting the critical role of the INR module in ensuring both privacy protection and reliable results.


\begin{figure}[htbp]
  \centering
  \includegraphics[width=1\textwidth]{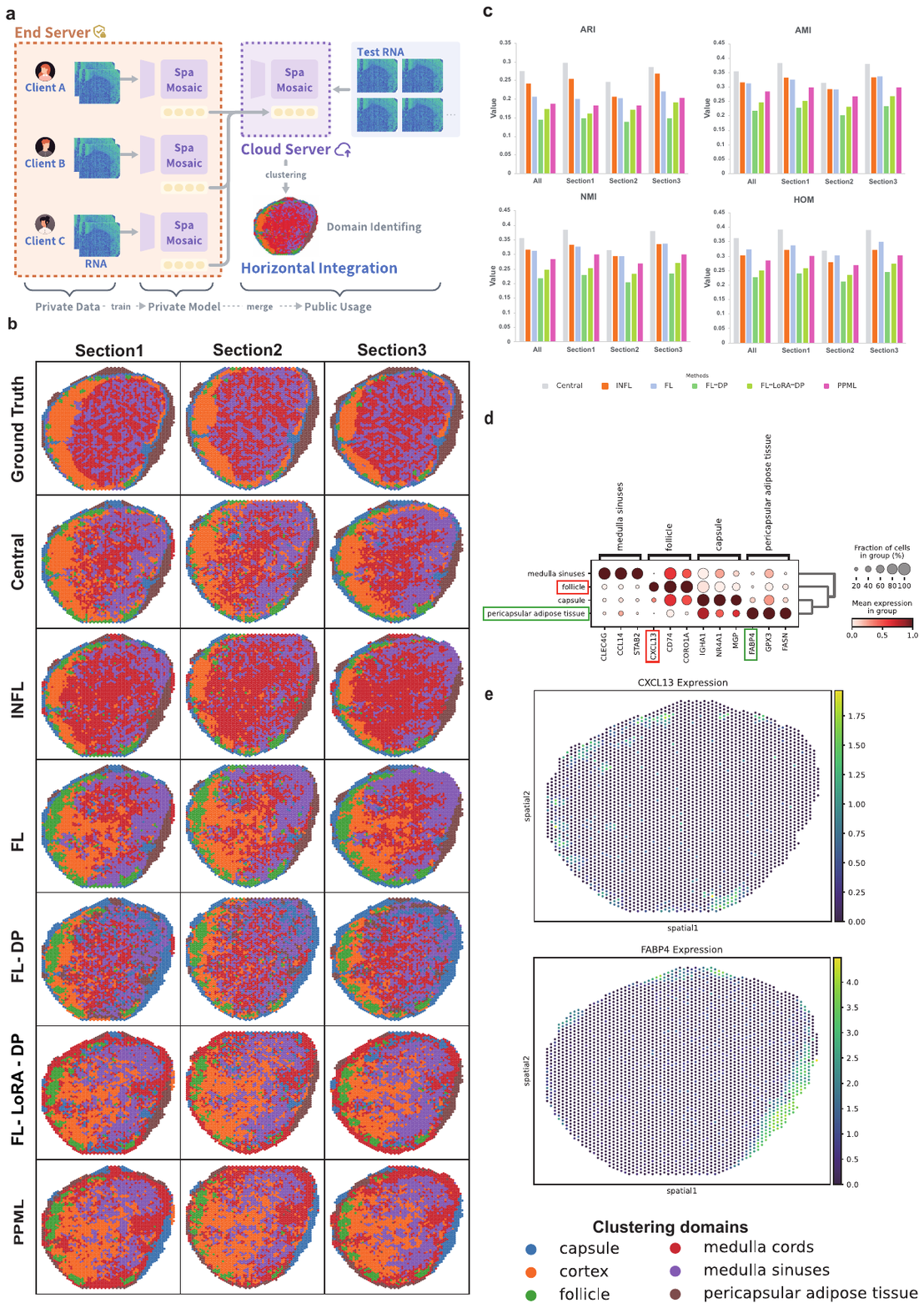}
    \caption{\textbf{Experimental results on the horizontal spatial transcriptomics integration task.}
\textbf{a} Experimental design of the algorithm framework for the spatial transcriptomics horizontal integration task, including training data partitioning, the training workflow, the final inference process and downstream clustering tasks.
\textbf{b} Spatial visualization of the clustering results.
\textbf{c} Quantitative metrics for the clustering performance, using four indices: ARI, AMI, NMI, and HOM.
\textbf{d} Differential expression analysis performed using the INFL clustering results, showing highly differentially expressed genes for each domain. We selected the genes \textit{CXCL13} and \textit{FABP4}, referenced relevant literature to confirm their biological significance and plotted their spatial distributions.
\textbf{e} Spatial distribution maps of the genes \textit{CXCL13} and \textit{FABP4}. Comparison with the INFL clustering map in panel \textbf{b} shows a strong spatial concordance between the expression of these genes and the locations of the follicle and pericapsular adipose tissue domains respectively, confirming the specificity of gene expression.
  }
    \label{fig:figure4}
\end{figure}

\subsection{INFL Excels in Horizontal Integration Tasks for Spatial Transcriptomics}\label{sec:results-Multi-Omics Integration}

With the development of omics technologies, spatial omics techniques are being widely applied in clinical research~\cite{walker2024toward}. An important task is to integrate data from different patient samples to explore common and differential features between patients. However, the privacy of clinical data makes it difficult to aggregate all spatial omics data from different sources for integrated training. Here, we demonstrate INFL as a federated learning strategy could yield an integration model while preserving privacy. 

We first explore the horizontal integration scenario of single omics, namely integrating the same omics data from different samples together and generating informative latent embeddings for each sample while considering inter-sample differences. As shown in \textbf{Figure~\ref{fig:figure4}a}, specifically, we selected three spatial RNA sections from the human lymph node dataset~\cite{yan2024mosaic}. We assumed that they belong to three different individuals' data. Simultaneously, we set up three clients, assuming they belong to three different hospitals. Then, we assigned two of the three RNA sections to each client at random, so that each client contains different data. We adopted the SpaMosaic model as our centralized algorithm, which is an efficient spatial omics integration algorithm in a horizontal setting~\cite{yan2024mosaic}. We applied various federated learning strategies to the baseline algorithm and initiated federated learning training. All clients participated in every communication round, including using local data for model training, uploading model parameters to the center, and updating the model based on the parameters returned from the center to continue training. Finally, the embeddings were generated using the globally aggregated model and all available RNA sections. The embeddings of each section were then clustered by leiden algorithm to distinguish spatial regions. The clustering labels were compared with the true labels to demonstrate the clustering performance of different federated learning strategies.

As shown in \textbf{Figure~\ref{fig:figure4}b}, we first plot the spatial distribution of clustering labels to qualitatively demonstrate the clustering effect. The INFL strategy performs on par with the centralized algorithm, accurately detecting the scattered distribution of follicles and the enveloping distribution of cortex, which is also consistent with the ground truth. INFL performs far better than other federated learning strategies (FL, FL-DP, FL-LoRA-DP, PPML). Other algorithms perform poorly in clustering and identifying regions, either failing to restore the enveloping distribution of cortex or accurately identify follicles. Our conclusion can also be obtained from quantitative metrics (\textbf{Figure~\ref{fig:figure4}c}), especially on the ARI index, where INFL outperforms the other algorithms and closely approaches the centralized algorithm.

To further investigate whether the INFL clustering labels have biological significance, we used the INFL clustering results to find marker genes for each domain and plotted the dotplot as shown in \textbf{Figure~\ref{fig:figure4}d}. From the dotplot, it can be seen that each domain can find its significantly expressed marker genes. \textit{CXCL13} is specifically expressed in the follicle region, which is consistent with literature records. \textit{CXCL13} is a chemokine secreted by follicular dendritic cells (FDC) and follicular reticular cells (B-cell zone reticular cells) within lymph node follicles. It binds to the receptor CXCR5 on the surface of B cells, guiding B cells to migrate to lymph node follicles, promoting follicle formation and maintenance~\cite{cosgrove2020b}. In addition, \textit{FABP4} is mainly expressed by mature adipocytes and macrophages. In pericapsular adipose tissue, \textit{FABP4} acts as a lipid chaperone protein, regulating fatty acid storage, transport, and metabolism, and also involved in inflammatory response regulation~\cite{liu2022fabp4}. Again, a significant positive correlation between this gene and domain can be observed in the dotplot. Furthermore, we displayed the spatial distribution of case genes (\textbf{Figure~\ref{fig:figure4}e}), \textit{CXCL13} is actually highly expressed in the follicle region, and \textit{FABP4} is no exception, highly expressed in pericapsular adipose tissue. These results further confirms that INFL clustering results can provide certain biological significance and shows the special spatial distribution of specific genes.

\begin{figure}[htbp]
  \centering
  \includegraphics[width=1\textwidth]{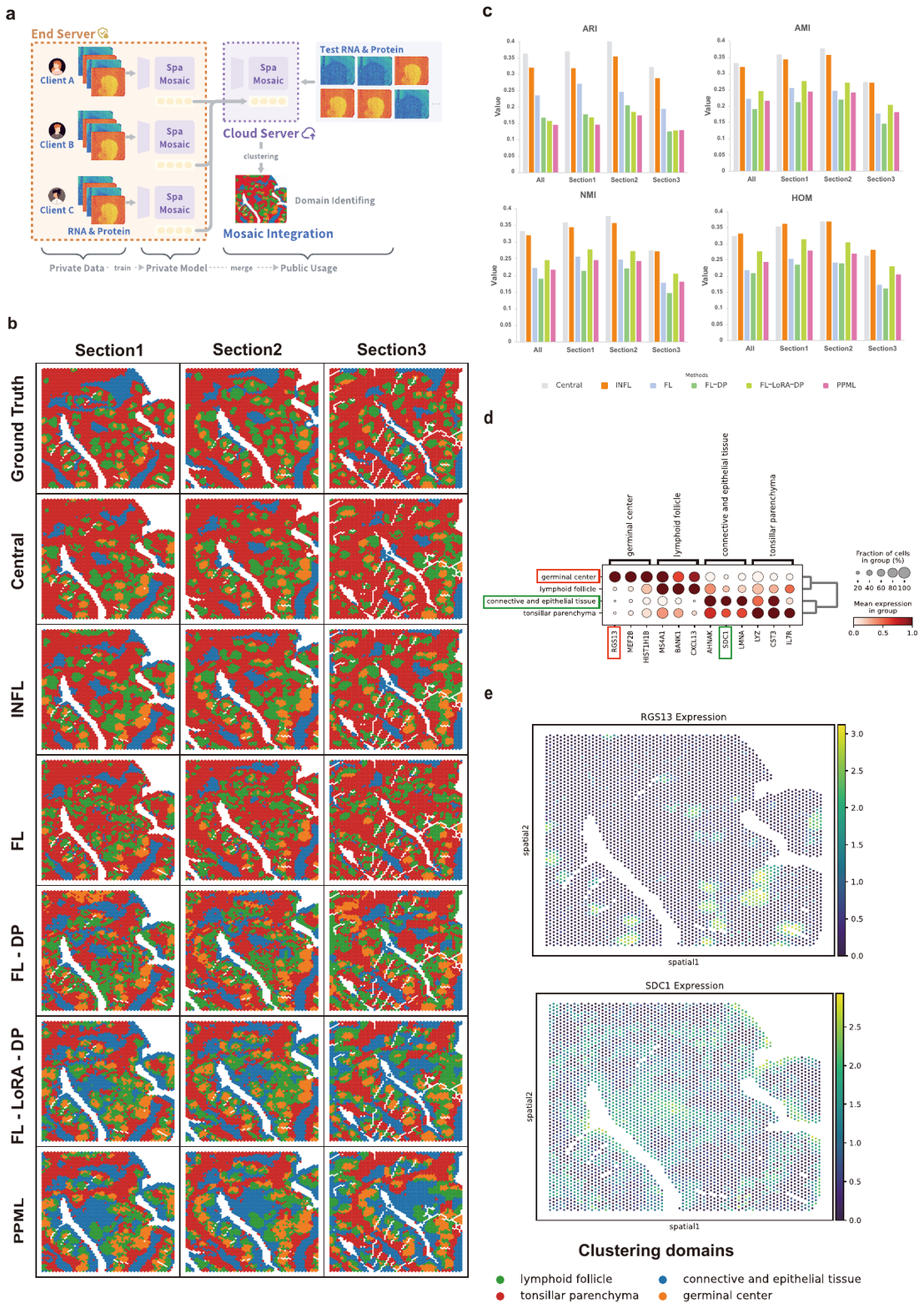}
    \caption{\textbf{Experimental results on the mosaic spatial multi-omics integration task.}
\textbf{a} Experimental design of the algorithm framework for the spatial multi-omics mosaic integration task, including training data partitioning, the training workflow, the final inference process, and downstream clustering tasks.
\textbf{b} Spatial visualization of the clustering results.
\textbf{c} Quantitative metrics for the clustering performance, using four indices: ARI, AMI, NMI, and HOM.
\textbf{d} Differential expression analysis performed using the INFL clustering results, showing highly differentially expressed genes for each domain. We selected the genes \textit{RGS13} and \textit{SDC1}, referenced relevant literature to confirm their biological significance, and plotted their spatial distributions.
\textbf{e} Spatial distribution maps of the genes \textit{RGS13} and \textit{SDC1}. Comparison with the INFL clustering map in panel \textbf{b} shows a strong spatial concordance between the expression of these genes and the locations of the germinal center and connective and epithelial tissue domains respectively, confirming the specificity of gene expression.
  }
    \label{fig:figure5}
\end{figure}

\subsection{INFL Excels in Mosaic Integration Tasks for Spatial Multi Omics}\label{sec:results-Multi-Omics Integration-Mosaic}

Beyond single omics integration, we further verify the effectiveness of INFL on multi-omics scenario, i.e., with simultaneous measurement of two or more types of molecules, by testing spatial multi-omics mosaic integration. As shown in \textbf{Figure~\ref{fig:figure5}a}, we used public data from the human tonsil dataset~\cite{yan2024mosaic}, containing three sections, each with two modalities (RNA and protein). We assumed that they belong to three different individuals' data. The mosaic integration scenario refers to the training setting where some sections have paired multi-omics data while others possess only single-omics data, aiming to generate informative latent embeddings for each sample while accounting for both inter-sample differences and multimodal information. We also set up three clients and assumed they belong to three different hospitals. Then we assigned mosaic data to each client. For example, for Client 1, we assigned paired RNA and protein modality data from Section 1, RNA single-modality data from Section 2, and protein single-modality data from Section 3. The data assigned to each client were different. We employed the SpaMosaic model as our centralized algorithm under the mosaic setting. Various federated learning strategies were applied to the baseline algorithm. All clients participated in every communication round, which included training the local models using local data, uploading local model parameters to the central server, and updating the local models based on the parameters returned from the central server to continue training. Finally, the embeddings were generated using the globally aggregated model and all available sections' multi-modality data. The embeddings of each section were then clustered using the leiden algorithm to distinguish spatial domains. The clustering labels were compared with the ground truth labels to evaluate the clustering performance of different federated learning strategies.

As shown in \textbf{Figure~\ref{fig:figure5}b}, we first plotted the spatial distribution of the clustering labels to qualitatively demonstrate the clustering performance. The INFL strategy performed on par with the centralized algorithm, accurately detecting lymphoid follicles surrounding the germinal center, as well as connective and epithelial tissue interwoven with tonsillar parenchyma, which is also consistent with the ground truth. The performance of INFL was significantly better than that of other federated learning strategies (FL, FL-DP, FL-LoRA-DP, PPML). Other algorithms showed poor performance in identifying spatial domains. For example, the federated learning method resulted in fragmented identification of lymphoid follicles with substantial noise, while PPML misidentified connective and epithelial tissue. Quantitative metrics similarly demonstrated this trend (Figure~\ref{fig:figure5}c), where INFL showed almost no gap compared to the centralized algorithm but outperformed other algorithms by approximately 5\% – 10\%.

Next, to investigate whether the clustering labels generated by INFL hold biological significance, we used the INFL clustering results to identify marker features for each domain. First, we generated a dot plot as shown in \textbf{Figure~\ref{fig:figure5}d}. The dot plot reveals that each domain has its significantly expressed marker genes. For instance, \textit{RGS13} was observed to be enriched in the germinal center region. Literature documents that this gene is a key marker for Germinal Center B cells in tonsillar tissue, closely associated with B cell activation, proliferation, and differentiation~\cite{yu2025parallel}. \textit{SDC1} is primarily expressed on the surface of epithelial cells, where it binds to collagen, fibronectin, laminin, and other components in the extracellular matrix (ECM) via its extracellular domain, forming physical connections between epithelial cells and connective tissue~\cite{kumar2021nuclear}. It plays a critical role in the connective and epithelial tissue region. Furthermore, we displayed the spatial distribution of case genes (\textbf{Figure~\ref{fig:figure5}e}), showing that \textit{RGS13} is indeed highly expressed in the germinal center region and \textit{SDC1} showed marked enrichment in its corresponding domain. Beyond genes, we also presented differential protein signals in \textbf{Extended Data Figure~\ref{fig:figure5-extend}}. The CXCR5 protein plays a central role in lymphoid follicle positioning, germinal center zoning, and the core signaling axis of T-B cell collaboration~\cite{wu2019cxcr4}. The figure shows that CXCR5 is differentially and significantly expressed in the lymphoid follicle and germinal center regions. This further confirms that the clustering results from INFL can provide biological insights, revealing specific spatial distributions of particular genes and proteins.

\section{Discussion}\label{sec:discussion}
Our proposed INFL demonstrates broad applicability across diverse biomedical scenarios, offering a comprehensive and scalable solution for privacy-preserving distributed learning. INFL supports tasks spanning from single-cell transcriptomics to spatial multi-omics, addressing challenges in cohort-scale classification, regression for perturbation prediction, and clustering in high-dimensional biological data. This versatility enables INFL to effectively handle heterogeneous and sensitive datasets across a variety of omics modalities, bridging the gap between privacy-preserving methodology and real-world biomedical applications. By ensuring robust performance in a diverse set of tasks, bulk proteomics classification, single-cell regression, and spatial transcriptomics clustering, INFL provides a unified method that meets the demands of modern biomedical research.

Ablation studies further underscore the critical role of the INR module in INFL. The INR module not only enhances model performance but also provides a robust mechanism for privacy protection. Specifically, without the INR module, the model experiences a significant drop in utility, which inadvertently reduces the effectiveness of potential attacks such as data reconstruction or misuse. Furthermore, unauthorized users lacking the correct INR key are unable to provide valid inputs for the coordinate-based encryption scheme, causing the model to output corrupted and meaningless results. This ensures that even if model parameters are exposed, sensitive data and intellectual property remain fully protected, as the model becomes entirely non-functional without the correct INR key. 

Moreover, INFL aligns well with the principles of Swarm Learning (SL), a decentralized framework that enables collaborative training across multiple sites without requiring data centralization. Similar to FL, INFL ensures data sovereignty by keeping sensitive datasets localized while allowing the exchange of learned parameters including the meta learner. This compatibility opens new avenues for applying INFL in scenarios where SL has been demonstrated to be successful, such as molecular classification in cancer histopathology and multi-modal clinical data integration~\cite{warnat2021swarm, saldanha2022swarm}. By integrating INR modules within SL, INFL could further enhance privacy and scalability, enabling robust federated learning across geographically dispersed biomedical institutions. This synergy has the potential to democratize access to high-quality AI models by fostering cross-institutional collaborations, ultimately accelerating innovation in biomedical research while adhering to strict privacy regulations.

Like all federated learning methods, INFL faces inherent trade-offs between privacy and performance. While federated learning typically sacrifices some accuracy compared to centralized training due to decentralized data and privacy constraints, our INR modules significantly mitigate this limitation by acting as meta-learners, enhancing generalization across heterogeneous datasets. In many cases, INFL achieves performance comparable to centralized training, as demonstrated in various omics tasks, while simultaneously providing strong privacy guarantees. However, challenges such as communication overhead and reliance on consistent client participation remain areas for improvement. Future research can focus on refining the INR architecture and further reducing computational costs. Despite these challenges, INFL represents a significant step forward in privacy-preserving federated learning, offering a practical, lightweight, and effective solution to drive scalable and secure biomedical AI.

\section{Method}\label{sec:method}
\subsection{Private Data Collection for Validation}
To address the challenges of data heterogeneity and privacy in biomedical research, we collected private datasets specifically tailored to evaluate the performance of our model on gene perturbation tasks. These private datasets were collected to simulate real-world experimental conditions while ensuring compliance with privacy regulations. They enable the systematic assessment of our method in modeling perturbation effects, including gene knockdowns and their impact on cellular phenotypes, critical for studying complex biological processes and validating the robustness of privacy-preserving INFL.

\textbf{Cell cultures:} Human primary umbilical vascular endothelial cells (HUVECs) were isolated and cultured as previously described~\cite{han2017zyxin}. Human embryonic kidney (HEK293T) cells were cultured at 37°C in a humidified atmosphere with 5\% CO2. The culture medium consisted of Dulbecco's Modified Eagle’s Medium (DMEM, C11995500BT, Invitrogen) supplemented with 10\% fetal bovine serum (FBS). 

\textbf{Virus preparation and infection:} A commercial lentiviral system (Sigma) was used to silence three 14-3-3 isoforms encoding genes, including YWHAB, YWHAE and YWHAH. The target and control scrambled sequences were selected from the human shRNA library of Sigma (http://www.sigmaaldrich.com/china-mainland/zh/life-science/functionalgenomics- and-rnai/SHRNA/library-information.html). 

Preparations of lentivirus were made in 293T cells. Briefly, 293T cells were seeded at a density of $1 \times 10^{4} cells/cm^2$ in 2 mL of complete growth medium and incubated for 24 hours until they reached 80–90\% confluency. Transfection complexes were prepared by combining polyethylenimine (PEI) with the target and packaging plasmids in serum-free medium at room temperature. The complexes were then added drop-wise to the cell culture. After 5 hours of incubation, the transfection medium was replaced with fresh complete medium, and the cells were further incubated for 48 hours. Subsequently, the virus-containing supernatant was collected.

For infection, HUVECs were incubated with a 1:1 mixture of viral supernatant and fresh medium, supplemented with polybrene at a final concentration of 8 µg/mL. After 24 hours, the virus-containing medium was aspirated and replaced with standard culture medium.

\textbf{Single-cell RNA sequencing:} Cells were harvested at 7 days post-infection for single-cell RNA sequencing (scRNA-seq). Prior to sequencing, the knockdown efficiency of the target gene was preliminarily validated by RT-qPCR. scRNA-seq was subsequently performed on both control and YWHA-knockdown groups, with all library preparation and sequencing services provided by Shanghai Biotechnology Corporation (Shanghai, China). Briefly, single-cell libraries were prepared using the 10x Genomics platform, where GEMs were formed to partition individual cells, enabling reverse transcription of mRNA into cDNA tagged with cell-specific 10x Barcodes and transcript-specific UMIs.



\subsection{Preprocessing for the Gene Perturbation}

Besides our private datasets, we employed two additional public datasets. First, we utilized the combinatorial CRISPR screening dataset generated in 2019~\cite{norman2019exploring} (referred as the Norman dataset) to comprehensively evaluate the performance of the INFL on gene perturbation task. This dataset, generated using the Perturb-seq technology, profiles approximately 200,000 human K562 cells (a leukemia cell line), covering single-gene and dual-gene knockdown perturbations targeting genes essential for cell growth and differentiation. Additionally, we incorporated the dataset released in 2016 ~\cite{adamson2016multiplexed} (referred as the Adamson dataset), which used Perturb-seq to profile CRISPR-mediated perturbations in the mammalian unfolded protein response (UPR). By combining single-cell RNA-seq with CRISPR barcoding, this study analyzed about 100 candidate genes identified from CRISPRi screens for their role in ER homeostasis, revealing functional gene clusters, the decoupling of UPR branches, and bifurcated activation states within the same perturbation.
The combination of all three datasets provides abundant training samples for the model to learn the complex, non-linear effects of genetic perturbations.

We applied specialized preprocessing methods tailored to different data components for each dataset, including the Norman, Adamson, and our private dataset. For the single-cell gene expression count matrix, we first split the expression data into perturbed and control cell populations. The control population consists of cells that received non-targeting guide RNAs, representing the basal expression state of the cells without perturbation. We normalized the total counts for each cell to match the median of total counts across all cells, and the normalized values were subsequently $log$-transformed with an offset of 1.

In addition to the expression matrix, the gene perturbation model requires a gene regulatory graph as prior knowledge inputs. We constructed this graph based on the Gene Ontology (GO) database~\cite{gene2004gene}. In this graph, each gene is represented as a node, and an edge is established between two genes if they co-belong to the same GO term. This process yields an undirected adjacency matrix $G$, where $G_{ij} = 1$ if gene $i$ and gene $j$ are functionally related, and 0 otherwise. The matrix $G$ is symmetric and includes self-loops for all nodes, i.e., $G_{ii} = 1$.

\subsection{Preprocessing for the Spatial Multi-omics Integration}

For the spatial proteome-transcriptome multi-omics integration task, we utilized two datasets generated by the 10x Genomics Visium platform for spatial RNA and protein co-profiling~\cite{kiessling2024spatial}: a human tonsil dataset and a human lymph node dataset. These datasets, derived from immune-related tissues, offer valuable insights into the human immune microenvironment. To achieve comprehensive integration, we performed both horizontal integration, aligning modalities within the same sections to correct for technical variability, and mosaic integration, bridging complementary modalities across different sections to construct a unified multi-omics representation.

We applied distinct preprocessing methods tailored to different modalities. For the spatial transcriptomics count matrix, we normalized the total count of each spot to match the median total counts across all spots. The normalized values were subsequently $log$-transformed with an offset of 1. We then selected the top 3,000 highly variable genes (HVGs) to serve as input to the model~\cite{yip2019evaluation}.

For the spatial proteomics count matrix, we performed the centered log ratio (CLR) transformation across spots, as defined by:
\begin{equation}
\text{CLR}(\mathbf{p}) = \left( \log\left(\frac{p_1}{g(\mathbf{p})}\right), \dots, \log\left(\frac{p_n}{g(\mathbf{p})}\right) \right)
\end{equation}
where $\mathbf{p} = (p_1, \dots, p_n)$ represents the count vector of protein epitopes for each spot, and $g(\mathbf{p})$ denotes the geometric mean of vector $\mathbf{p}$.

In addition to the count matrices, the model requires a spatial adjacency graph as input. We constructed a k-nearest neighbors (kNN) graph based on the spatial coordinates of the spots, with each spot represented as a node, while the edges were derived from the Euclidean distances between the spots \cite{danielsson1980euclidean}. The parameter \(k\) was set to 10 by default for the tonsil data, while for the human lymphoid dataset, it was adjusted to 2 to minimize over-smoothing. This yielded an undirected adjacency matrix $A$, where $A_{ij} = 1$ if spot $i$ is among the $k$ nearest neighbors of spot $j$, and 0 otherwise. The matrix $A$ is symmetric and includes self-loops, i.e., $A_{ii} = 1$.

\subsection{Preprocessing for Proteomic Analysis}

Proteomic data were collected exclusively from Cohort 1 of the ProCan Compendium, consisting of 766 tumor samples and 494 tumor-adjacent normal samples, derived from a total of 638 patients~\cite{cai2025federated}. All biospecimens were obtained from pathology laboratories and biobanks, including the Victorian Cancer Biobank, the Gynaecological Oncology Biobank (GynBiobank) at Westmead Hospital, and the Children’s Medical Research Institute Legacy sample set in Australia. Ethical approvals for the collection and use of these samples were obtained prior to analysis.

The raw DIA-MS data were processed into a protein quantification matrix, where rows represent samples and columns correspond to protein abundances. Proteins detected in fewer than $50\%$ of samples were excluded. Missing values were imputed with zeros, and protein abundance values were log$_2$-transformed. No cohort-level normalization was applied to avoid introducing biases into the analysis.

Samples were filtered based on strict quality control criteria. Only those with technical replicate correlations above 0.9 were included to ensure data reliability. Samples that failed histopathologic validation, exhibited tumor content below 20\%, or had necrosis levels exceeding 80\% were excluded from downstream analyses. In addition, data from Hepatoblastoma, Retinoblastoma, Ganglioneuroblastoma, and Rhabdoid Tumor samples were excluded from the analysis due to their limited sample size, which was insufficient to allow practical train–test split.

\subsection{Design of Computational Experiments}

\textbf{Baselines:} We evaluated INFL against the traditional FedAvg algorithm~\cite{mcmahan2017communication} and three representative privacy-preserving federated learning (FL) baseline methods, described as follows:

\begin{itemize}
    \item \textbf{FL}: The traditional FedAvg algorithm~\cite{mcmahan2017communication}, which aggregates client parameters on a central server through simple averaging. This serves as the foundational approach in federated learning.
    
    \item \textbf{FL-DP}: A privacy-preserving variant of the FedAvg algorithm that incorporates differential privacy (DP)~\cite{dwork2006differential}. This method enhances the privacy of federated learning by adding noise to the model updates, mitigating the risk of exposing sensitive client information.
    
    \item \textbf{FL-LoRA-DP}: An extension of FL-DP that integrates low-rank adaptation (LoRA) modules~\cite{hu2022lora} into the local models. By introducing LoRA, this approach not only preserves privacy through DP but also improves model performance by fine-tuning model representations in resource-efficient ways.
    
    \item \textbf{PPML}: A more advanced privacy-preserved PPML-Omics~\cite{zhou2024ppml} building upon FL-LoRA-DP. It incorporates a decentralized randomization (DR) algorithm~\cite{massoulie2007randomized} to further enhance privacy by applying randomized transformations at a decentralized level, offering stronger protection while maintaining model accuracy.
\end{itemize}

These baselines provide a comprehensive evaluation setup, encompassing both traditional and state-of-the-art privacy-preserving techniques in federated learning.

\textbf{Data Distribution and Federated Scenarios}: We evaluated the INFL under three distinct data distribution scenarios:
\begin{itemize}
    \item \textbf{INFL-Proteomic}: The dataset was partitioned into five groups. Each of the five clients was assigned one group.
    \item \textbf{INFL-Perturbation}: The dataset was partitioned into ten equal, non-overlapping shards. Each of the ten clients was assigned one shard.
    \item \textbf{INFL-SpaIntegration-Horizontal}: Three clients were instantiated. From a dataset comprising three RNA data slices, each client was randomly assigned two of the three slices. This simulates the scenario where collaborators have partially overlapping datasets.
    \item \textbf{INFL-SpaIntegration-Mosaic}: To simulate a more complex, realistic scenario of non-IID and multi-modal data overlap, we distributed data from three slices (Slice 1, 2, 3) with two modalities (RNA, ADT) among three clients as follows:
    \begin{itemize}
        \item \textbf{Client 1}: RNA from Slices 1 \& 2; ADT from Slices 1 \& 3.
        \item \textbf{Client 2}: RNA from Slices 1 \& 2; ADT from Slices 2 \& 3.
        \item \textbf{Client 3}: RNA from Slices 2 \& 3; ADT from Slices 1 \& 3.
    \end{itemize}
    This configuration ensures no client possesses the complete dataset, making federated collaboration essential.
\end{itemize}

\textbf{Implementation Details and Hyperparameters:} All models were implemented using Python with the PyTorch and PyTorch Geometric libraries and trained on NVIDIA GeForce RTX 3090 GPUs. It is worth noting that INFL can be applied in two scenarios: parameter-efficient fine-tuning for traditional vision tasks using Vision Transformers, as detailed in the supplementary material, and model pretraining, as discussed in the main manuscript.
\begin{itemize}
    \item For \textbf{INFL-Proteomic}, training was conducted for 200 global rounds with 5 clients, a client participation fraction of 1.0, and 50 local epochs per round. The Adam optimizer was used with a learning rate of $1 \times 10^{-4}$. The hidden size of the INR module is set to 8 with 3 layers.
    \item For \textbf{INFL-Perturbation}, training was conducted for 200 global rounds with 10 clients, a client participation fraction of 0.5, and 2 local epochs per round. The Adam optimizer was used with a learning rate of $5 \times 10^{-3}$. The hidden size of the INR module is set to 8 with 3 layers.
    \item For \textbf{INFL-SpaIntegration}, training ran for 50 global rounds with 3 clients, a client participation fraction of 1.0, and 2 local epochs per round. The Adam optimizer was used with a learning rate of $10^{-3}$. The hidden size of the INR module is set to 8 with 3 layers.
    \item For \textbf{DP settings}, we used a gradient clipping norm of $C=1.0$, a target delta of $\delta=10^{-5}$, and noise multipliers of $\sigma=1.29$ (DP-series) and $\sigma=2.36$ (PPML)~\cite{zhou2024ppml}.
    \item For the \textbf{LoRA settings}, we set the rank to 4 and the scaling parameter $\alpha$ to 1.0. The LoRA weights are initialized using Kaiming initialization~\cite{he2015delving}, which ensures stable training by adjusting the variance of the weights based on the number of input units.
\end{itemize}

\subsection{Implicit Neural Federated Learning Protocol and Model Encryption}

To enable collaborative model development on sensitive, decentralized datasets, we developed Implicit Neural Federated Learning (INFL), a method that integrates a privacy-preserving training protocol with a robust mechanism for intellectual property (IP) protection. For the privacy-preserving component, INFL utilizes the Federated Averaging (\texttt{FedAvg}) algorithm to train a global model across a network of clients without centralizing their private data. The process is organized into discrete global communication rounds, indexed by $t \in \{1, \dots, T\}$, and proceeds as follows:

\begin{enumerate}
    \item \textbf{Initialization}: The central server initializes a global model, $M_G^0$, with randomly initialized parameters.
    \item \textbf{Client Selection}: At the start of each round $t$, the server selects a subset of clients, $S_t$, comprising a fraction $C$ of the total $K$ available clients.
    \item \textbf{Model Distribution}: The server transmits the current global model parameters, $M_G^{t-1}$, to all selected clients $k \in S_t$.
    \item \textbf{Local Training}: Each client $k$ performs local training on its private dataset $D_k$ for $E$ epochs. The client updates the model parameters by minimizing a task-specific objective function using stochastic gradient descent, yielding a local model update $M_k^t$.
    \item \textbf{Model Aggregation}: All participating clients in $S_t$ transmit their updated parameters $M_k^t$ to the server. The server then aggregates these local models to compute the new global model $M_G^t$ by averaging their parameters:
    \begin{equation}
    M_G^t = \frac{1}{|S_t|} \sum_{k \in S_t} M_k^t
    \end{equation}
\end{enumerate}

While the federated protocol protects the training data, the resulting global model itself constitutes valuable IP. To secure it against unauthorized use, we developed a novel, coordinate-based encryption scheme built upon Implicit Neural Representations (INRs). This method embeds a secret key directly into the model's architecture, rendering the parameters non-functional without it. The core of this approach is the replacement of standard fully-connected layers (\texttt{nn.Linear}) in the model's decoder with a custom \texttt{INRLinear} module. A standard linear layer computes its output as $\mathbf{y} = \mathbf{W}\mathbf{x} + \mathbf{b}$. Our \texttt{INRLinear} module modifies this by incorporating a modulation term, $\mathbf{\Delta} \in \mathbb{R}^{N_{\text{out}}}$, generated by a small MLP known as the INR network, $\Phi_{\theta}$. The final layer output $\mathbf{y}'$ is a balanced combination:
\begin{equation}
\mathbf{y}' = \alpha(\mathbf{W}\mathbf{x} + \mathbf{b}) + (1-\alpha)\mathbf{\Delta}
\end{equation}
where $\alpha$ is a hyperparameter. The encryption is embedded in the generation of $\mathbf{\Delta}$. We introduce a secret key, $\pi$, which is a random permutation of the output indices $\{0, 1, \dots, N_{\text{out}}-1\}$. For each output neuron $i$, its coordinate $c'_i$ is derived from this secret permutation, not its natural index:
\begin{equation}
c'_i = \frac{2\pi(i)}{N_{\text{out}}-1} - 1, \quad \text{for } i \in \{0, 1, \dots, N_{\text{out}}-1\}
\end{equation}
These key-dependent coordinates are then mapped to a high-dimensional feature space using a sinusoidal positional encoding function, $\gamma(\cdot)$, to capture high-frequency details:
\begin{equation}
\gamma(c) = \left( \sin(2^0\pi c), \cos(2^0\pi c), \dots, \sin(2^{L-1}\pi c), \cos(2^{L-1}\pi c) \right)
\end{equation}
The INR network $\Phi_{\theta}$ takes these encoded coordinates as input to produce the modulation for each neuron: $\mathbf{\Delta}_i = \Phi_{\theta}(\gamma(c'_i))$. During training, the entire model learns to perform its task conditioned on the coordinate system defined by $\pi$. An authorized user, possessing both the model parameters and the key, can generate the correct modulation $\mathbf{\Delta}$ for valid predictions. Conversely, an attacker with only the parameters would be forced to use a default permutation, feeding scrambled coordinates to the INR, resulting in a meaningless modulation and a corrupted, non-functional model output. Thus, INFL achieves both data privacy during training and IP protection for the final model.

\section{INFL Model Architectures}

\textbf{INFL-Perturbation Model Architecture:} The foundation for our perturbation prediction experiments is based on a Graph Neural Network (GNN) designed to predict transcriptomic responses to genetic perturbations. The model, which uses a hidden dimension of 64, utilizes two separate GNNs. First, a 1-layer Simple Graph Convolution (SGC) network operates on a gene co-expression graph to learn positional gene embeddings. This graph is constructed by connecting genes with a Pearson correlation greater than 0.4, keeping up to 20 neighbors per gene. Second, a parallel 1-layer SGC network processes a Gene Ontology (GO) similarity graph (connecting up to 20 most similar genes) to learn embeddings for perturbations. To predict the outcome of a specific perturbation, the model adds the learned perturbation embedding to the basal gene embeddings. This combined representation is then processed by a hierarchical decoder, which comprises a shared MLP and two subsequent gene-specific linear layers, to generate the final post-perturbation expression vector. The model is trained by minimizing the discrepancy between prediction and ground-truth expression profiles.

\textbf{INFL-SpaIntegration Model Architecture:} For spatial transcriptomics integration, we employed a multi-level graph construction strategy and a Weighted Graph Convolutional Network (WLGCN) encoder. Intra-modality graphs are built by connecting spots within a spatial radius of 2000 units or based on the 10 nearest neighbors. To bridge different modalities and batches, the model identifies the 10 mutual nearest neighbors (MNNs) in the PCA-reduced expression space, assigning these inter-graph edges a weight of 0.8.

The core of the model is the WLGCN, which acts as an encoder. The forward pass proceeds as follows: (1) Principal Component Analysis (PCA)-reduced expression profiles serve as initial node features. (2) A series of 8 WLGCN layers perform message passing, correctly accounting for edge weights to differentiate between spatial and MNN connections. The outputs from the initial features and all subsequent layers are concatenated to form a multi-scale feature representation. (3) This representation is passed through a fully connected layer projecting to a hidden dimension of 512, followed by batch normalization, a LeakyReLU activation with a negative slope of 0.2, and dropout with a probability of 0.2 to produce a low-dimensional latent embedding. (4) The final embedding is L2-normalized for use in downstream analyses. A parallel decoder module, configured by default as a single linear layer, reconstructs the initial node features from this latent embedding, enabling the computation of a reconstruction loss for model training.

\textbf{INFL-Proteomic Model Architecture:} The foundation of our proteomic classification framework is built on a neural network model designed to classify proteomic profiles into distinct cancer subtypes. The model employs a modular architecture that incorporates embedding, residual learning, and hierarchical decoding layers to capture the high-dimensional and complex nature of proteomic data~\cite{cai2025federated}. 

The model begins with an embedding layer, a linear transformation that projects the input proteomic features into a hidden feature space of dimension 64. This representation is then passed through two sequential residual blocks, each designed to enhance feature extraction while preserving the original input information. Within each residual block, two fully connected layers are interleaved with non-linear activation functions (\texttt{ReLU}), batch normalization (or layer normalization, depending on the variant), and dropout layers to prevent overfitting. A residual connection is applied to ensure stable gradient propagation, with the output of each block normalized and summed with the input.

The output of the second residual block is further processed by a hierarchical decoder, consisting of two fully connected layers. The first layer reduces the dimensionality of the learned features to 32, followed by a second layer that outputs predictions corresponding to the number of target classes 15. The decoder incorporates dropout regularization and a non-linear activation (\texttt{ReLU}) to ensure robust feature transformation and classification.

The model is trained by minimizing the cross-entropy loss between the predicted and true class labels, using the Adam optimizer for parameter updates. The architecture's modular design, with residual connections and hierarchical decoding, ensures effective learning from high-dimensional proteomic data while maintaining stability during training.


\subsection{Evaluation and Metrics}
Model performance was assessed using a combination of metrics tailored to each task. For the classification tasks, we adopted the metrics below to evaluation the performance:  
\begin{itemize}
    \item \textbf{Classification Accuracy}:  
    The proportion of correctly classified samples among all test samples, measuring the overall predictive accuracy of the model across all cancer subtypes. It is defined as:  
    \[
    \text{Accuracy} = \frac{\text{Number of Correct Predictions}}{\text{Total Number of Samples}}
    \]  

    \item \textbf{F1-Score}:  
    The harmonic mean of precision and recall, computed for each cancer subtype and averaged (macro-F1) to assess the model’s ability to balance false positives and false negatives across all classes. For each class, the F1-score is defined as:  
    \[
    \text{F1} = 2 \cdot \frac{\text{Precision} \cdot \text{Recall}}{\text{Precision} + \text{Recall}}
    \]  
    where:  
    \[
    \text{Precision} = \frac{\text{True Positives}}{\text{True Positives} + \text{False Positives}}, \quad     \]  
    \[ \text{Recall} = \frac{\text{True Positives}}{\text{True Positives} + \text{False Negatives}}
    \]  

    \item \textbf{Area Under the Receiver Operating Characteristic Curve (AUC)}:  
    The AUC is calculated for each cancer subtype in a one-vs-rest setting and averaged (macro-AUC) to evaluate the model's ability to discriminate between classes based on predicted probabilities.  

    \item \textbf{Macro-Averaged AUROC}:  
    The average of the AUROC values computed for each cancer subtype in a one-vs-rest setup, providing an overall measure of the model’s ability to distinguish between all cancer subtypes. It is defined as:  
    \[
    \text{Macro-Averaged AUROC} = \frac{1}{C} \sum_{c=1}^{C} \text{AUROC}_c
    \]  
    where \(C\) is the total number of cancer subtypes, and \(\text{AUROC}_c\) is the AUROC for each class \(c\).

    \item \textbf{SHAP (SHapley Additive exPlanations) Values}:  

    SHAP values quantify the contribution of each protein feature to individual cancer type predictions by fairly distributing the difference between the model's output and expected baseline across all features. Based on cooperative game theory, SHAP values satisfy desirable properties including efficiency, symmetry, dummy feature, and additivity. It is calculated as:
    
    \[ 
    \phi_j(f, x) = \sum_{S \subseteq F \setminus \{j\}} \frac{|S|!(|F|-|S|-1)!}{|F|!} [f(S \cup \{j\}) - f(S)]
    \]
    
    where $\phi_j(f, x)$ is the SHAP value for protein feature $j$, $F$ is the set of all protein features, $S$ is a subset of features excluding $j$, $f(S)$ is the model's expected output given feature subset $S$, and the sum is over all possible subsets $S$ of features not including $j$.
    
\end{itemize}

For the prediction tasks (i.e., INFL-Perturbation), we need to evaluate the predictive performance, where the following metrics are used:

\begin{itemize}
    \item \textbf{Overall Mean Squared Error (MSE)}:  
    The MSE between predicted ($\hat{y}_g$) and true ($y_g$) expression values across all $G$ genes, measuring global prediction accuracy. It is defined as:  
    \[
    \text{MSE} = \frac{1}{G} \sum_{g=1}^{G} (\hat{y}_g - y_g)^2
    \]

    \item \textbf{MSE on DE genes with opposite direction of change ($\text{MSE}_{\text{de}}$)}:  
    The MSE computed exclusively on the top 20 most differentially expressed (DE) genes for a given perturbation where the predicted expression changes in the \textbf{opposite direction} of the true change. This metric assesses the model's ability to avoid incorrect directional predictions for key biological signals. It is defined as:  
    \[
    \text{MSE}_{\text{de}} = \frac{1}{|G_{\text{opp}}|} \sum_{g \in G_{\text{opp}}} (\hat{y}_g - y_g)^2
    \]
    where \(G_{\text{opp}}\) is the set of top 20 DE genes for which the predicted change is in the \textbf{opposite direction} of the true change, i.e., \((\hat{y}_g - y_g) \cdot y_g < 0\), and \(|G_{\text{opp}}|\) is the number of such genes.

    \item \textbf{Fraction of top 20 differentially expressed genes with opposite prediction direction (De\_op\_frac)}: 
    The percentage of the top 20 most differentially expressed genes that are predicted to change expression in the opposite direction compared to the true direction of change. This metric identifies cases where the model not only fails to predict the correct magnitude but also predicts the wrong regulatory direction. It is calculated as:
    \[ 
    \text{De\_op\_frac} = \frac{1}{N} \sum_{i=1}^{N} \frac{1}{20} \sum_{j \in \text{Top20}_{DE}} \mathbb{I}[\text{sign}(\Delta y_{ij}) \neq \text{sign}(\Delta \hat{y}_{ij})]
    \]
    where $\Delta y_{ij}$ and $\Delta \hat{y}_{ij}$ are the true and predicted expression changes for gene $j$ in sample $i$, and $\mathbb{I}[\cdot]$ is the indicator function.
    
    \item \textbf{Overall Pearson Correlation ($\rho$)}:  
    The Pearson correlation between the vectors of mean predicted and mean true gene expression, measuring the preservation of linear relationships in the expression profiles. It is defined as:  
    \[
    \rho = \frac{\sum_{g=1}^{G} (\hat{y}_g - \bar{\hat{y}})(y_g - \bar{y})}{\sqrt{\sum_{g=1}^{G} (\hat{y}_g - \bar{\hat{y}})^2} \sqrt{\sum_{g=1}^{G} (y_g - \bar{y})^2}}
    \]  
    where $\bar{\hat{y}}$ and $\bar{y}$ are the mean predicted and true expression values, respectively.

    \item \textbf{Pearson correlation on DE genes with correct direction of change ($\text{Pearson}_{\text{de}}$)}:  
        The Pearson correlation coefficient computed on the top 20 most differentially expressed (DE) genes for a given perturbation where the predicted expression changes in the **same direction** as the true change. This metric assesses the model's ability to capture the correct trend of biological signals. It is defined as:  
        \[
        \text{Pearson}_{\text{de}} = \frac{\text{Cov}(y_g, \hat{y}_g)}{\sigma_{y_g} \sigma_{\hat{y}_g}}
        \]
        where:
        \begin{itemize}
        \item  \(G_{\text{correct}}\) is the set of top 20 DE genes for which the predicted change is in the \textbf{same direction} as the true change, i.e., \((\hat{y}_g - y_g) \cdot y_g > 0\).
        \item \(\text{Cov}(y_g, \hat{y}_g)\) is the covariance between the true expression values (\(y_g\)) and predicted values (\(\hat{y}_g\)) for genes in \(G_{\text{correct}}\).
        \item \(\sigma_{y_g}\) and \(\sigma_{\hat{y}_g}\) are the standard deviations of the true and predicted expression values, respectively, for genes in \(G_{\text{correct}}\).
        \end{itemize}

    \item \textbf{Pearson correlation on Delta Expression ($\text{Pearson}_{\Delta}$)}:  
        The Pearson correlation coefficient computed between the **predicted change** in post-perturbation gene expression relative to the unperturbed control expression and the **true change** in post-perturbation gene expression relative to the control. This metric evaluates the model's ability to capture the relative changes in gene expression induced by perturbations. It is defined as:  
        \[
        \text{Pearson}_{delta} = \frac{\text{Cov}(\Delta \hat{y}_g, \Delta y_g)}{\sigma_{\Delta \hat{y}_g} \sigma_{\Delta y_g}}
        \]
        where:
        \begin{itemize}
        \item  \(\Delta \hat{y}_g = \hat{y}_g - \hat{y}_{g,\text{control}}\) is the predicted change in gene expression for gene \(g\) after perturbation relative to the control.
        \item \(\Delta y_g = y_g - y_{g,\text{control}}\) is the true change in gene expression for gene \(g\) after perturbation relative to the control.
        \item \(\text{Cov}(\Delta \hat{y}_g, \Delta y_g)\) is the covariance between predicted and true changes in gene expression.
        \item \(\sigma_{\Delta \hat{y}_g}\) and \(\sigma_{\Delta y_g}\) are the standard deviations of the predicted and true changes, respectively.
        \end{itemize}
\end{itemize}

For the task where the effectiveness of data integration is the key (i.e., INFL-SpaIntegration), we adopt the following evaluation metrics for computing data integration efficiency:
\begin{itemize}

    \item \textbf{Biological Structure Preservation (ARI)}:  
    The Adjusted Rand Index (ARI) compares the clustering of the integrated embeddings ($\mathcal{C}$) against ground-truth cell type labels ($\mathcal{T}$), where an ARI score close to 1 signifies excellent preservation. It is defined as:  
    \[
    \text{ARI} = \frac{\sum_{i,j} \binom{n_{ij}}{2} - \left[ \sum_{i} \binom{a_{i}}{2} \sum_{j} \binom{b_{j}}{2} \right] / \binom{n}{2}}{\frac{1}{2} \left[ \sum_{i} \binom{a_{i}}{2} + \sum_{j} \binom{b_{j}}{2} \right] - \left[ \sum_{i} \binom{a_{i}}{2} \sum_{j} \binom{b_{j}}{2} \right] / \binom{n}{2}}
    \]  
    where $n_{ij}$ is the number of samples in both cluster $i$ and ground-truth $j$, $a_i$ is the size of cluster $i$, $b_j$ is the size of ground-truth $j$, and $n$ is the total number of samples.

    \item \textbf{Normalized Mutual Information (NMI)}:  
    The Normalized Mutual Information (NMI) quantifies the agreement between the clustering of integrated embeddings ($\mathcal{C}$) and ground-truth labels ($\mathcal{T}$). NMI is normalized to a range of [0, 1], where 1 indicates perfect alignment. It is defined as:  
    \[
    \text{NMI} = \frac{2 \cdot I(\mathcal{C}; \mathcal{T})}{H(\mathcal{C}) + H(\mathcal{T})}
    \]  
    where $I(\mathcal{C}; \mathcal{T})$ is the mutual information between $\mathcal{C}$ and $\mathcal{T}$, and $H(\mathcal{C})$ and $H(\mathcal{T})$ are the entropies of $\mathcal{C}$ and $\mathcal{T}$, respectively.

    \item \textbf{Adjusted Mutual Information (AMI)}:  
    The Adjusted Mutual Information (AMI) adjusts the raw mutual information to correct for chance, providing a more robust comparison of clustering quality. Like NMI, AMI ranges from 0 to 1, with higher values indicating better alignment. It is defined as:  
    \[
    \text{AMI} = \frac{I(\mathcal{C}; \mathcal{T}) - \mathbb{E}[I(\mathcal{C}; \mathcal{T})]}{\max(H(\mathcal{C}), H(\mathcal{T})) - \mathbb{E}[I(\mathcal{C}; \mathcal{T})]}
    \]  
    where $\mathbb{E}[I(\mathcal{C}; \mathcal{T})]$ is the expected mutual information under random clustering.

    \item \textbf{Batch Effect Removal (HOM)}:  
    The Homogeneity (HOM) metric quantifies the ability to remove batch effects while maintaining biological signal. HOM is calculated using the entropy of batch label distributions within clusters. A lower HOM score indicates better batch mixing, while preserving biological structure. It is defined as:  
    \[
    \text{HOM} = - \frac{1}{C} \sum_{c=1}^{C} \sum_{b=1}^{B} p_{cb} \log(p_{cb})
    \]  
    where $C$ is the number of clusters, $B$ is the number of batches, and $p_{cb}$ is the proportion of batch $b$ samples in cluster $c$. A lower HOM score reflects better batch effect removal with minimal disruption of biological structure.

\end{itemize}

\subsection{Biological Analysis and Interpretation of Model Outputs}
\textbf{Analysis of INFL-Proteomic:} To interpret the contributions of individual protein features to the model's predictions, we employed SHapley Additive exPlanations (SHAP), a game-theory-based framework for explaining machine learning model outputs. Specifically, we used the \texttt{shap.GradientExplainer}, which leverages the model's gradient information to efficiently approximate Shapley values for each input feature. The input feature matrix, composed of protein expression data, was passed through a trained federated learning model in evaluation mode. SHAP values were computed to quantify the marginal contribution of each protein feature to the predicted post-perturbation cancer type classification relative to a control baseline. This analysis enabled the identification of key protein features driving the model's predictions and provided insights into their biological relevance. To interpret the classification results for each cancer type, we selected the top five proteins with the greatest contributions and visualized them using beeswarm plots. This allowed us to elucidate the magnitude and direction (positive or negative correlation) of the association between each protein and the cancer classification. We also searched literature evidence to support our observation results.

\textbf{Analysis of INFL-Perturbation:} We used the \texttt{gears.plot\_perturbation} function to visualize the actual expression changes (displayed as box plots) of the top 20 differential genes under a specific perturbation condition, along with the mean change in gene expression predicted by different methods. To quantitatively characterize the prediction performance, we introduced the hit rate metric. A prediction is considered accurate if the mean change in gene expression predicted by a method falls within the range of the box plot; the overall prediction accuracy for the 20 genes is then calculated as the hit rate. This can be expressed by the formula:

\[
\text{Hit rate} = \frac{\text{Number of correctly predicted genes}}{\text{Total number of genes (generally 20)}}
\]

\textbf{Analysis of INFL-SpaIntergration:} After model integration to obtain the embeddings, we performed clustering using Leiden algorithm, implemented via the \texttt{Scanpy} package~\cite{wolf2018scanpy}, and annotated each domain with a name. By comparing clustering results from different methods overlaid with the ground truth on the same canvas, the qualitative accuracy of clustering was assessed. We used the \texttt{sc.pl.embedding} function to visualize the clustering results. To reveal the biological significance of the clusters, we identified marker genes for each domain using \texttt{sc.rank\_genes\_groups}, visualized their spatial distributions, and validated whether these marker genes are indeed specifically expressed in the corresponding regions with support from the literature.


\subsection{Theoretical Analysis of the INR Encryption Mechanism}
\label{sec:theoretical_analysis}

We provide a formal analysis to demonstrate that the \texttt{INRLinear} module, conditioned on a secret permutation key $\pi$, functions as a cryptographic lock. We show that without knowledge of $\pi$, the model's output is mathematically equivalent to adding a high-variance, structured noise term, rendering the model non-functional.

\begin{definition}[Model Output]
Let an \texttt{INRLinear} layer be parameterized by its base weights $\mathbf{W}, \mathbf{b}$, the INR network $\Phi_{\theta}$, and a secret permutation key $\pi$ over the indices $\{0, 1, \dots, N_{\text{out}}-1\}$. For an input $\mathbf{x}$, the output $\mathbf{y}' \in \mathbb{R}^{N_{\text{out}}}$ is given by:
\begin{equation}
    \mathbf{y}'_i = \alpha(\mathbf{W}\mathbf{x} + \mathbf{b})_i + (1-\alpha)\mathbf{\Delta}_i
\end{equation}
where the modulation term $\mathbf{\Delta}_i$ is generated using the secret key $\pi$:
\begin{equation}
    \mathbf{\Delta}_i = \Phi_{\theta}\left(\gamma\left(\frac{2\pi(i)}{N_{\text{out}}-1} - 1\right)\right)
\end{equation}
\end{definition}

\begin{proposition}[Unauthorized Access]
An unauthorized party possesses the model parameters $(\mathbf{W}, \mathbf{b}, \theta)$ but lacks the secret key $\pi$. The most rational strategy for the attacker is to assume a default, or identity, permutation, $\pi_A(i) = i$.
\end{proposition}

Under this assumption, the attacker computes an output, which we denote $\mathbf{y}'_A$, using $\pi_A$. The attacker's computed modulation term, $\mathbf{\Delta}_A$, is therefore:
\begin{equation}
    (\mathbf{\Delta}_A)_i = \Phi_{\theta}\left(\gamma\left(\frac{2\pi_A(i)}{N_{\text{out}}-1} - 1\right)\right) = \Phi_{\theta}\left(\gamma\left(\frac{2i}{N_{\text{out}}-1} - 1\right)\right)
\end{equation}
The resulting output computed by the attacker is:
\begin{equation}
    (\mathbf{y}'_A)_i = \alpha(\mathbf{W}\mathbf{x} + \mathbf{b})_i + (1-\alpha)(\mathbf{\Delta}_A)_i
\end{equation}

To quantify the functional difference, we analyze the error vector $\boldsymbol{\epsilon} = \mathbf{y}' - \mathbf{y}'_A$. The error for the $i$-th output neuron is:
\begin{align}
    \epsilon_i &= (\mathbf{y}'_i) - (\mathbf{y}'_A)_i \\
    &= \left[ \alpha(\mathbf{W}\mathbf{x} + \mathbf{b})_i + (1-\alpha)\mathbf{\Delta}_i \right] - \left[ \alpha(\mathbf{W}\mathbf{x} + \mathbf{b})_i + (1-\alpha)(\mathbf{\Delta}_A)_i \right] \\
    &= (1-\alpha) \left( \mathbf{\Delta}_i - (\mathbf{\Delta}_A)_i \right)
\end{align}
The core of the encryption lies in the relationship between the true modulation $\mathbf{\Delta}$ and the attacker's computed modulation $\mathbf{\Delta}_A$. Let us examine the components. The true modulation for neuron $i$ uses the coordinate derived from $\pi(i)$. The attacker's modulation for neuron $i$ uses the coordinate derived from $i$.

We can define a mapping for the coordinate generation from $C(j) = \frac{2j}{N_{\text{out}}-1} - 1$:
\begin{align}
    \mathbf{\Delta}_i &= \Phi_{\theta}(\gamma(C(\pi(i)))) \\
    (\mathbf{\Delta}_A)_i &= \Phi_{\theta}(\gamma(C(i)))
\end{align}
Notice that the set of all coordinates used is the same for both the authorized user and the attacker: $\{C(0), C(1), \dots, C(N_{\text{out}}-1)\}$. However, the key $\pi$ shuffles the assignment of these coordinates to the output neurons.

We can define a new vector $\mathbf{\Delta}^*$ whose $j$-th component is the modulation generated by the $j$-th canonical coordinate:
\begin{equation}
    \mathbf{\Delta}^*_j = \Phi_{\theta}(\gamma(C(j)))
\end{equation}
Using this definition, we can express both $\mathbf{\Delta}$ and $\mathbf{\Delta}_A$ more clearly:
\begin{align}
    \mathbf{\Delta}_i &= \mathbf{\Delta}^*_{\pi(i)} \\
    (\mathbf{\Delta}_A)_i &= \mathbf{\Delta}^*_{i}
\end{align}
This reveals that the attacker's modulation vector $\mathbf{\Delta}_A$ is simply the canonical modulation vector $\mathbf{\Delta}^*$. The authorized user's modulation vector $\mathbf{\Delta}$ is a permutation of $\mathbf{\Delta}^*$ (or equivalently, of $\mathbf{\Delta}_A$) according to the secret key $\pi$. Specifically, $\mathbf{\Delta}$ is obtained by applying the permutation $\pi$ to the indices of the elements of $\mathbf{\Delta}_A$.

Substituting this back into the error term $\epsilon_i$:
\begin{equation}
    \epsilon_i = (1-\alpha) \left( \mathbf{\Delta}^*_{\pi(i)} - \mathbf{\Delta}^*_{i} \right)
\end{equation}
The error at output $i$ is proportional to the difference between the modulation value that \emph{should} be at index $\pi(i)$ and the modulation value that \emph{should} be at index $i$. Since $\pi$ is a random permutation, for a large $N_{\text{out}}$, $\pi(i) \neq i$ with high probability. The INR network $\Phi_{\theta}$, combined with the sinusoidal positional encoding $\gamma$, is trained to be a high-frequency function that maps specific coordinates to specific modulation values required for the task. The values in $\mathbf{\Delta}^*$ are therefore highly structured and non-uniform. The term $\mathbf{\Delta}^*_{\pi(i)} - \mathbf{\Delta}^*_{i}$ represents the difference between two distinct, quasi-randomly chosen components from the learned modulation vector.

Assuming the components of $\mathbf{\Delta}^*$ are learned values with zero mean ($\mathbb{E}[\mathbf{\Delta}^*_j] = 0$) and variance $\sigma^2_{\Delta}$, the expected value of the error at any given neuron is:
\begin{equation}
    \mathbb{E}[\epsilon_i] = (1-\alpha) \left( \mathbb{E}[\mathbf{\Delta}^*_{\pi(i)}] - \mathbb{E}[\mathbf{\Delta}^*_{i}] \right) = 0
\end{equation}
However, the expected squared error (variance) is substantial. Assuming $\mathbf{\Delta}^*_{\pi(i)}$ and $\mathbf{\Delta}^*_{i}$ are uncorrelated for $i \neq \pi(i)$:
\begin{align}
    \mathbb{E}[\epsilon_i^2] &= (1-\alpha)^2 \mathbb{E}[(\mathbf{\Delta}^*_{\pi(i)} - \mathbf{\Delta}^*_{i})^2] \\
    &= (1-\alpha)^2 \left( \mathbb{E}[(\mathbf{\Delta}^*_{\pi(i)})^2] - 2\mathbb{E}[\mathbf{\Delta}^*_{\pi(i)}\mathbf{\Delta}^*_{i}] + \mathbb{E}[(\mathbf{\Delta}^*_{i})^2] \right) \\
    &\approx (1-\alpha)^2 (\sigma^2_{\Delta} + \sigma^2_{\Delta}) = 2(1-\alpha)^2 \sigma^2_{\Delta}
\end{align}
This demonstrates that the attacker's output $\mathbf{y}'_A$ is corrupted by an additive, zero-mean error term with a variance proportional to the variance of the learned modulation signal. This error term is not random noise but a deterministic scrambling of the necessary correction signal. The scrambling operation, dictated by the unknown permutation $\pi$, effectively decouples the corrective modulation from its intended neuronal target, corrupting the final output and rendering the model's predictions meaningless. Thus, the secret key $\pi$ acts as a functional lock on the model, providing robust IP protection.

\noindent\textbf{Extreme Case Analysis: Inference without the Modulation Network}
\label{sec:extreme_case_analysis}

We now analyze a more severe scenario of unauthorized access, where an attacker possesses the base model weights $(\mathbf{W}, \mathbf{b})$ but has no access to the parameters $\theta$ of the INR network $\Phi_{\theta}$. This represents a situation where the IP protection mechanism is completely removed. In this case, the attacker is unable to compute the modulation term $\mathbf{\Delta}$ at all. The most logical course of action for the attacker is to treat the \texttt{INRLinear} layer as a scaled standard linear layer, effectively setting the modulation term to zero.

\begin{proposition}[Inference without INR Parameters]
An unauthorized party possesses the base weights $(\mathbf{W}, \mathbf{b})$ but not the INR parameters $\theta$. The attacker's computed output for the layer, denoted $\mathbf{y}'_A$, is:
\begin{equation}
    (\mathbf{y}'_A)_i = \alpha(\mathbf{W}\mathbf{x} + \mathbf{b})_i
\end{equation}
This is because the term $(1-\alpha)\mathbf{\Delta}_i$ cannot be computed and is therefore omitted.
\end{proposition}

The correct output $\mathbf{y}'_i$, as computed by an authorized user with the key $\pi$ and the INR network $\Phi_{\theta}$, remains:
\begin{equation}
    \mathbf{y}'_i = \alpha(\mathbf{W}\mathbf{x} + \mathbf{b})_i + (1-\alpha)\mathbf{\Delta}_i
\end{equation}
where $\mathbf{\Delta}_i = \Phi_{\theta}(\gamma(C(\pi(i))))$.

We can now directly compute the error vector $\boldsymbol{\epsilon} = \mathbf{y}' - \mathbf{y}'_A$ resulting from the absence of the INR network. The error for the $i$-th output neuron is:
\begin{align}
    \epsilon_i &= (\mathbf{y}'_i) - (\mathbf{y}'_A)_i \\
    &= \left[ \alpha(\mathbf{W}\mathbf{x} + \mathbf{b})_i + (1-\alpha)\mathbf{\Delta}_i \right] - \left[ \alpha(\mathbf{W}\mathbf{x} + \mathbf{b})_i \right] \\
    &= (1-\alpha)\mathbf{\Delta}_i
    \label{eq:error_no_inr}
\end{align}

This result is critically important. The error in the attacker's computation is not a complex, scrambled signal as in the previous case; it is directly proportional to the modulation signal $\mathbf{\Delta}$ itself. This reveals the fundamental role of the INR during training. The optimization process does not treat the base linear transformation and the INR modulation as independent. Instead, it co-adapts them to solve the task jointly. The base weights $(\mathbf{W}, \mathbf{b})$ are trained to produce an "incomplete" or "biased" output, with the full expectation that the INR network will provide the necessary, precisely structured correction signal $\mathbf{\Delta}$ to arrive at the final, correct output.

Let the ideal, task-solving output of the layer be $\mathbf{y}_{\text{ideal}}$. During training, the entire layer is optimized such that $\mathbf{y}' \approx \mathbf{y}_{\text{ideal}}$. Therefore:
\begin{equation}
    \alpha(\mathbf{W}\mathbf{x} + \mathbf{b}) + (1-\alpha)\mathbf{\Delta} \approx \mathbf{y}_{\text{ideal}}
\end{equation}
The attacker, by computing only $\mathbf{y}'_A = \alpha(\mathbf{W}\mathbf{x} + \mathbf{b})$, is effectively computing:
\begin{equation}
    \mathbf{y}'_A \approx \mathbf{y}_{\text{ideal}} - (1-\alpha)\mathbf{\Delta}
\end{equation}
The attacker's output is systematically biased away from the ideal output by the exact amount of the missing corrective signal. The term $(1-\alpha)\mathbf{\Delta}$ is not random noise; it is a vital component of the learned function. Its absence guarantees a functional failure.

We can quantify the magnitude of this failure by analyzing the expected squared error. Using the canonical modulation vector $\mathbf{\Delta}^*$ defined previously, we have $\mathbf{\Delta}_i = \mathbf{\Delta}^*_{\pi(i)}$. The expected squared error at neuron $i$ is:
\begin{align}
    \mathbb{E}[\epsilon_i^2] &= \mathbb{E}[((1-\alpha)\mathbf{\Delta}_i)^2] \\
    &= (1-\alpha)^2 \mathbb{E}[(\mathbf{\Delta}^*_{\pi(i)})^2]
\end{align}
Assuming that the permutation $\pi$ is chosen uniformly at random, the expectation over the choice of $\pi$ for a given $i$ is equivalent to averaging over all possible indices $j \in \{0, \dots, N_{\text{out}}-1\}$. Let $\sigma^2_{\Delta}$ be the variance of the components of $\mathbf{\Delta}^*$ (assuming zero mean for simplicity).
\begin{equation}
    \mathbb{E}_{\pi}[\mathbb{E}[\epsilon_i^2]] = (1-\alpha)^2 \frac{1}{N_{\text{out}}} \sum_{j=0}^{N_{\text{out}}-1} (\mathbf{\Delta}^*_j)^2 = (1-\alpha)^2 \sigma^2_{\Delta}
\end{equation}
The resulting error has a substantial, non-zero variance that is directly dependent on the magnitude of the learned modulation signal. Since the model relies on this signal to function, $\sigma^2_{\Delta}$ will be significantly greater than zero. Therefore, completely removing the INR network is equivalent to removing a core computational block of the model, leading to a systematic and catastrophic collapse in performance. The base model on its own is not merely a degraded version of the full model; it is functionally incomplete and incapable of performing the learned task.

\textbf{Resistance to gradient-based attacks in federated training.}
\label{sec:attack}
Beyond the no-key inference settings, we further analyze the case where an adversary observes (or even controls) gradients during federated learning. Let a client hold a mini-batch $\mathcal{B}$ and consider one \texttt{INRLinear} layer with parameters $(\mathbf{W}, \mathbf{b}, \theta)$ and secret permutation $\pi$. For an input $\mathbf{x}$, the layer output is:
\begin{equation}
    \mathbf{y}'(\mathbf{x}) \;=\; \alpha(\mathbf{W}\mathbf{x}+\mathbf{b}) \;+\; (1-\alpha)\,\mathbf{\Delta}(\pi), 
    \quad 
    \mathbf{\Delta}_i(\pi)\;=\;\Phi_{\theta}\!\left(\gamma\!\left(C(\pi(i))\right)\right),
\end{equation}
with $C(j)=\tfrac{2j}{N_{\text{out}}-1}-1$. Let $\ell(\mathbf{x},\mathbf{t})$ be the sample loss and $L=\frac{1}{|\mathcal{B}|}\sum_{(\mathbf{x},\mathbf{t})\in\mathcal{B}}\ell(\mathbf{x},\mathbf{t})$ the batch loss.

We compare the true gradient under the secret key $\pi$ and the adversary's surrogate gradient computed under an identity-key assumption $\pi_A(i)=i$. Denote by $\mathbf{g}=\nabla_{\theta}L(\pi)$ and $\mathbf{g}_A=\nabla_{\theta}L(\pi_A)$ the corresponding gradients w.r.t.\ INR parameters, and similarly $\nabla_{\mathbf{W}}L(\pi)$ vs.\ $\nabla_{\mathbf{W}}L(\pi_A)$ for the base weights.

\textbf{Key observation (gradient permutation mismatch).}
By the chain rule,
\begin{equation}
    \frac{\partial L(\pi)}{\partial \theta}
    \;=\;
    \sum_{(\mathbf{x},\mathbf{t})\in\mathcal{B}}
    \sum_{i=0}^{N_{\text{out}}-1}
    \underbrace{\frac{\partial \ell}{\partial y'_i}(\mathbf{x},\mathbf{t})}_{\text{task gradient}}
    \cdot
    (1-\alpha)\;
    \underbrace{\frac{\partial \Phi_{\theta}(\gamma(C(\pi(i))))}{\partial \theta}}_{\text{INR Jacobian at }C(\pi(i))}.
    \label{eq:true_grad}
\end{equation}
An adversary who assumes $\pi_A$ computes instead
\begin{equation}
    \frac{\partial L(\pi_A)}{\partial \theta}
    \;=\;
    \sum_{(\mathbf{x},\mathbf{t})\in\mathcal{B}}
    \sum_{i=0}^{N_{\text{out}}-1}
    \frac{\partial \ell}{\partial \tilde{y}_i}(\mathbf{x},\mathbf{t})
    \cdot
    (1-\alpha)\;
    \frac{\partial \Phi_{\theta}(\gamma(C(i)))}{\partial \theta},
    \label{eq:adv_grad}
\end{equation}
where $\tilde{\mathbf{y}}$ is computed with $\pi_A$. The sets of coordinates $\{C(\pi(i))\}$ and $\{C(i)\}$ are identical, but their \emph{assignment} to output indices $i$ is permuted.

Assuming (i) $\mathbb{E}[\partial \ell/\partial y'_i]=0$ at stationarity and (ii) for $i\neq j$, the task gradients and INR Jacobians at different coordinates are weakly correlated (a standard assumption for high-frequency INR features), we obtain:
\begin{align}
    \mathbb{E}\!\left[\big\|\mathbf{g}-\mathbf{g}_A\big\|_2^2\right]
    &= (1-\alpha)^2\,
       \mathbb{E}\!\left[
           \Big\|
             \sum_{i}
               \frac{\partial \ell}{\partial y'_i}\,
               \Big(
                 \partial_{\theta}\Phi_{\theta}(\gamma(C(\pi(i))))
                 -
                 \partial_{\theta}\Phi_{\theta}(\gamma(C(i)))
               \Big)
           \Big\|_2^2
       \right]
       \nonumber\\
    &\approx (1-\alpha)^2 \sum_{i}
       \mathbb{E}\!\left[\Big(\tfrac{\partial \ell}{\partial y'_i}\Big)^2\right]\;
       \mathbb{E}\!\left[
         \left\|
           \partial_{\theta}\Phi_{\theta}(\gamma(C(\pi(i))))
           -
           \partial_{\theta}\Phi_{\theta}(\gamma(C(i)))
         \right\|_2^2
       \right].
       \label{eq:grad_mse}
\end{align}
Because $\pi$ is a random permutation, with high probability $\pi(i)\neq i$, and for high-frequency INRs the Jacobians at distinct coordinates behave like high-variance, weakly correlated features. Hence
\begin{equation}
    \mathbb{E}\!\left[
      \left\|
        \partial_{\theta}\Phi_{\theta}(\gamma(C(\pi(i))))
        -
        \partial_{\theta}\Phi_{\theta}(\gamma(C(i)))
      \right\|_2^2
    \right]
    \;\gtrsim\; 2\,\sigma_{\text{J}}^2,
\end{equation}
where $\sigma_{\text{J}}^2$ denotes the per-coordinate Jacobian variance. Substituting into \eqref{eq:grad_mse} yields
\begin{equation}
    \mathbb{E}\!\left[\big\|\mathbf{g}-\mathbf{g}_A\big\|_2^2\right]
    \;\gtrsim\;
    2(1-\alpha)^2\,\sigma_{\text{J}}^2 \sum_{i}\mathbb{E}\!\left[\Big(\tfrac{\partial \ell}{\partial y'_i}\Big)^2\right],
    \label{eq:grad_gap}
\end{equation}
which is strictly positive unless all task gradients vanish. Therefore, the adversary’s surrogate gradient is \emph{systematically biased} relative to the true gradient.

\textbf{Implications for gradient leakage and model inversion.}
Common federated attacks (e.g., gradient matching, model inversion) rely on the fidelity of observed gradients to the true per-example gradients. From \eqref{eq:true_grad}–\eqref{eq:grad_gap}, any gradient constructed using the wrong key (or assuming identity) replaces the correct per-neuron alignment with a permutation-mismatched one. Consequently:
\begin{itemize}
    \item The attacker’s per-example gradients are scrambled by an unknown permutation at the INR interface, breaking the one-to-one correspondence between output-residuals and INR coordinates required for gradient matching.
    \item The expected cosine similarity $\mathbb{E}[\langle \mathbf{g},\mathbf{g}_A\rangle / (\|\mathbf{g}\|\,\|\mathbf{g}_A\|)]$ is driven toward $0$ under weak-correlation assumptions, degrading reconstruction attacks.
    \item Even if the attacker updates $\theta$ using $\mathbf{g}_A$, the update direction deviates from the true descent direction, yielding optimization failure or convergence to a suboptimal, non-functional model.
\end{itemize}
An analogous argument holds for base weights $(\mathbf{W},\mathbf{b})$: via the chain rule, their gradients inherit the same permutation mismatch through the error signal backpropagated from the INR branch, yielding a non-vanishing gradient gap $\mathbb{E}\big[\|\nabla_{\mathbf{W}}L(\pi)-\nabla_{\mathbf{W}}L(\pi_A)\|^2\big]$ under the same conditions.

\subsection{Statistics and Reproducibility}
For all experiments, data were randomly selected from public or private datasets and no statistical methods were used to predetermine sample sizes. In addition, for the ease of the experiments, we leverage the coordinate inputs generated according to the shape of the weight tensor as the meta learners input. To ensure reproducibility, all model training processes were conducted with fixed random seeds (1 for INFL-Perturbation, 1234 for INFL-SpaIntegration, and 0 for Proteomic). All code and experimental configurations are available upon reasonable request.

Additionally, we conducted an ablation study to assess the privacy-preserving capabilities of INFL. Notably, when unauthorized attackers input random coordinates without the correct key, the model completely collapsed, outputting N/A and failing to generate any meaningful results. Furthermore, under a more extreme scenario where the INR module was removed during inference (\textbf{Figure~\ref{fig:figure2}g}), the model's accuracy dropped significantly, falling far below PPML and other baselines. These findings underscore the importance of INFL's key-conditioned mechanism in ensuring robust privacy protection while maintaining reliable performance. A detailed theoretical analysis is provided in Section~\ref{sec:extreme_case_analysis} to demonstrate how INFL retains its privacy-preserving capabilities under such extreme circumstances.

\section{Data Availability}\label{sec:data}

The data used in this work were partly obtained from the public dataset, downloaded according to the instructions provided in each corresponding reference, and partly collected by ourselves. We also upload the collection of the self-collected data for reproduction in \url{https://doi.org/10.6084/m9.figshare.30763670}.

\section{Code Availability}\label{sec:code}

Code to reproduce the experiments in this work or to compress your own data can be found at \url{https://github.com/RoyZry98/INFL-Pytorch}.

\section{Acknowledgements}\label{sec:acknowledgements}

\section{Author Contributions Statement}\label{sec:Author}
R.Z. designed the main methodology, performed the primary experiments, and wrote the original draft. H.D. and G.D. contributed to parts of the investigation, algorithm development, and writing. S.Z., Y.Z., A.C., and X.C. contributed to parts of the experiments. Z.Q., J.L., and P.L. assisted with data collection, curation and provided guidance on the biomedical application and manuscript editing. Y.D., X.X., J.C., and S.Z. supervised the research, acquired funding, administered the project, and were responsible for reviewing and editing the final manuscript. All authors reviewed and approved the final manuscript.

\section{Competing Interests Statement}\label{sec:Competing}
The authors declare no competing interests.

\renewcommand{\figurename}{Extended Data Fig.}

\begin{figure}[t]
\renewcommand{\thefigure}{1} 
  \centering
  \includegraphics[width=1\textwidth]{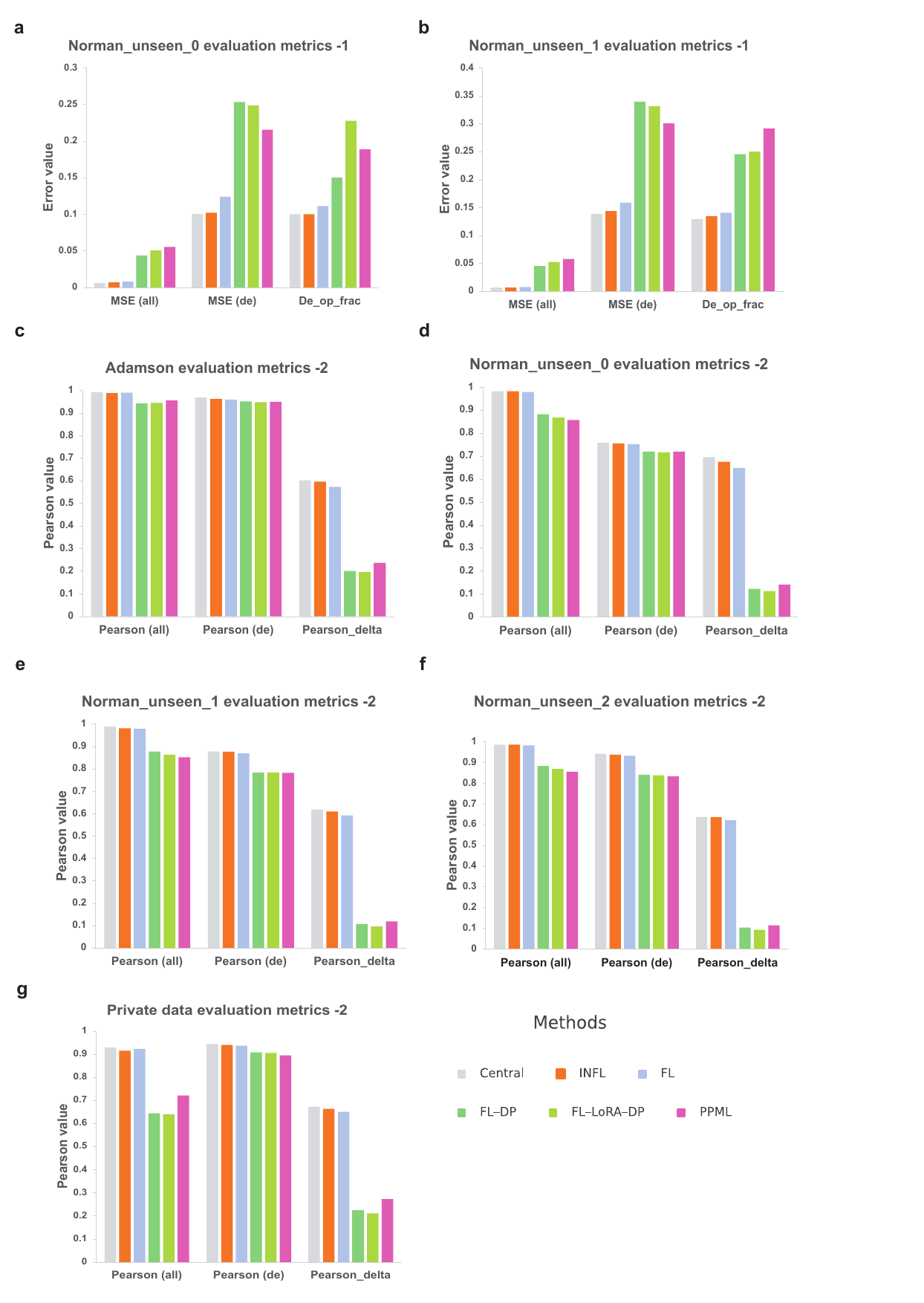}
    \caption{\textbf{Supplementary quantitative metrics for perturbation prediction performance.}
\textbf{a-b} Model perturbation prediction performance characterized by quantitative metrics. The metrics are the same as in Figure~\ref{fig:figure3}c-e, where lower values indicate better performance. \textbf{a} Performance on the Norman dataset under the in-domain (full) scenario. \textbf{b} Performance on the Norman dataset under the in-domain (partial) scenario.
\textbf{c-g} Model perturbation prediction performance characterized by other quantitative metrics. For each scenario, we selected three metrics to demonstrate the performance of each method: the Pearson correlation coefficient (PCC) for all genes (Pearson(all)), the PCC for the top 20 differentially expressed genes (Pearson(de)), and the Pearson\_delta (measuring the change). For all three metrics, higher values indicate better performance.
\textbf{c} Performance on the Adamson dataset. \textbf{d} Performance on the Norman dataset under the in-domain (full) scenario. \textbf{e} Performance on the Norman dataset under the in-domain (partial) scenario. \textbf{f} Performance on the Norman dataset under the out-of-domain scenario. \textbf{g} Performance on the private dataset.
  }
    \label{fig:figure3-extend-1}
\end{figure}

\begin{figure}[t]
\renewcommand{\thefigure}{2}
  \centering
  \includegraphics[width=1\textwidth]{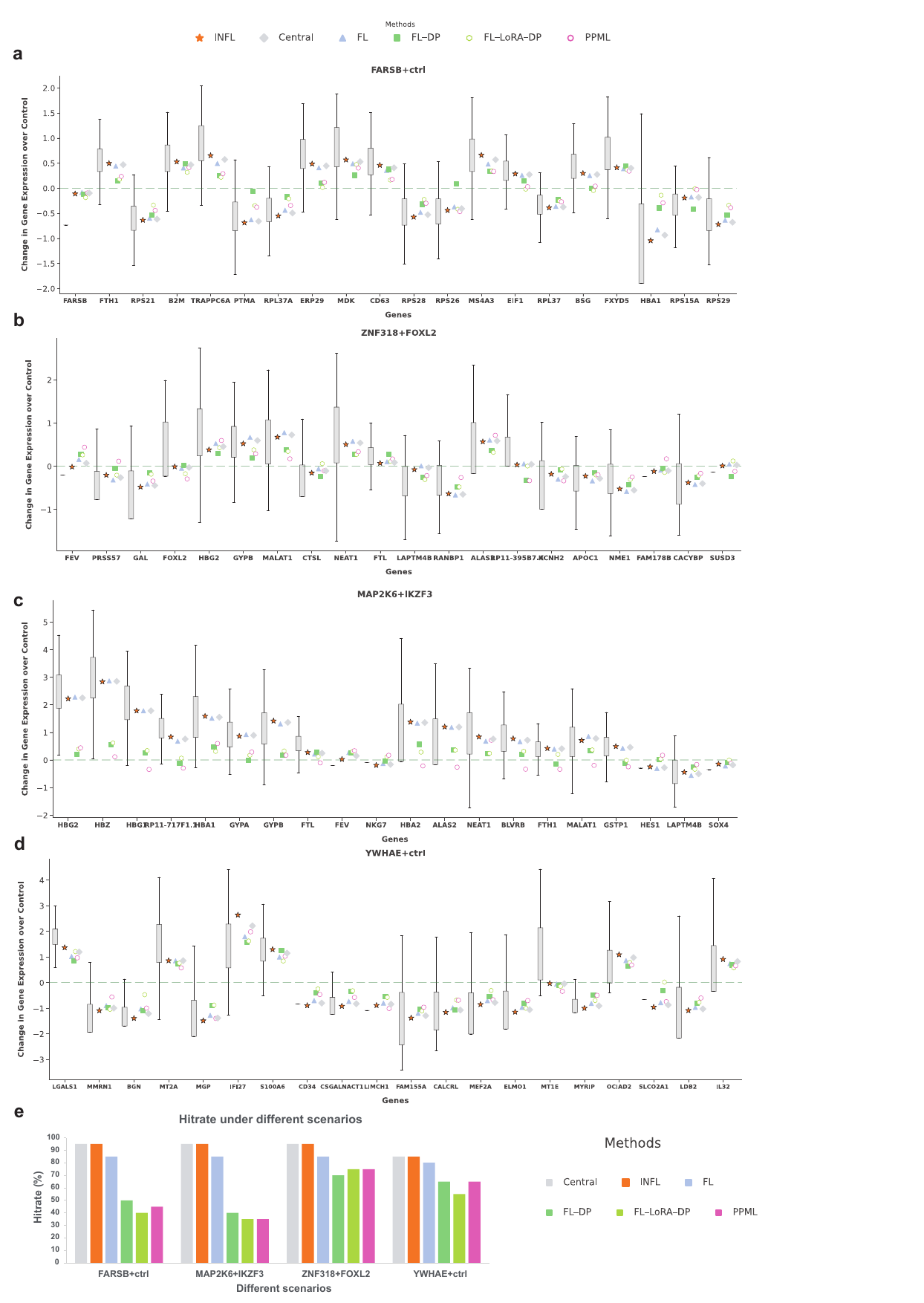}
    \caption{\textbf{Qualitative presentation of additional cases and hit rate analysis.}
\textbf{a-d} Qualitative case studies showing model perturbation performance. The green dotted line shows mean unperturbed control gene expression. Boxes indicate experimentally measured differential gene expression after perturbing some genes. Different symbols shows the mean change in gene expression predicted by different methods. Whiskers represent the last data point within 1.5× interquartile range.
\textbf{a} Case of \textit{FARSB} single-gene perturbation (from the Adamson dataset).
\textbf{b} Case of \textit{ZNF318} and \textit{FOXL2} combined perturbation (from the Norman dataset, under the in-domain (full) scenario).
\textbf{c} Case of \textit{MAP2K6} and \textit{IKZF3} combined perturbation (from the Norman dataset, under the in-domain (partial) scenario).
\textbf{d} Case of \textit{YWHAE} single-gene perturbation (from the private dataset).
\textbf{e} Perturbation performance for different cases quantitatively displayed using hit rate. 
  }
    \label{fig:figure3-extend-2}
\end{figure}

\begin{figure}[t]
\renewcommand{\thefigure}{3}
  \centering
  \includegraphics[width=\textwidth]{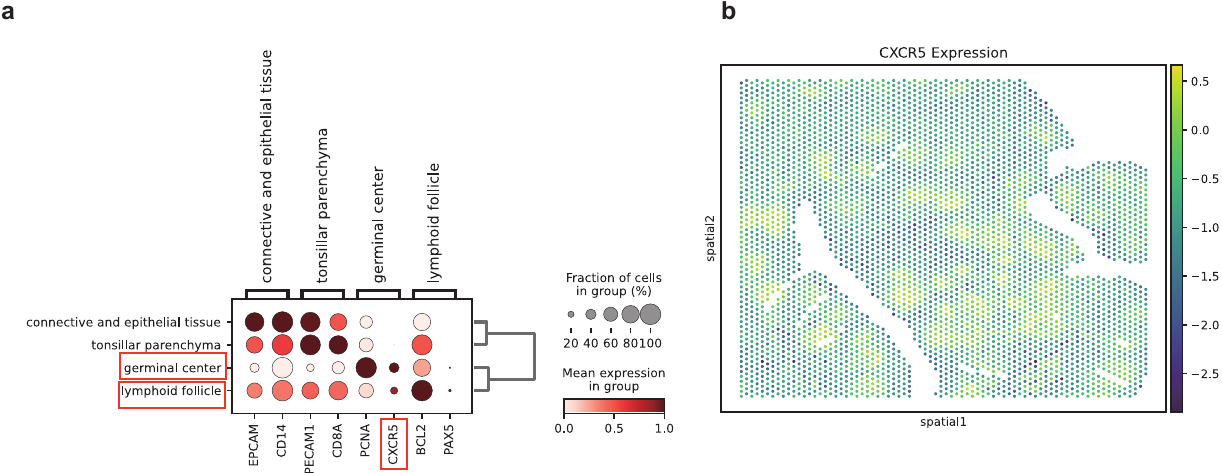}
  \caption{\textbf{Additional experimental results on protein validation for the mosaic integration task.}
  \textbf{a} Differential expression analysis based on the INFL clustering results, showing the most highly differentially expressed proteins for each domain. We selected the protein CXCR5, consulted relevant literature to confirm its biological significance, and plotted its spatial distribution.
  \textbf{b} Spatial distribution of the protein CXCR5. Compared with the INFL clustering map in Figure~\ref{fig:figure5}b, its expression shows strong spatial concordance with the germinal center and lymphoid follicle domains, confirming the specificity of this protein.}
  \label{fig:figure5-extend}
\end{figure}

\newpage
\begin{appendices}

\section{Supplementary Materials}
\subsection{Supplementary Figure 1}
\label{sec:sup}
We conducted a comprehensive evaluation of INFL on four canonical vision scenarios: (i) fine-grained classification on Stanford Cars and Stanford Dogs~\cite{krause2013stanfordcars,khosla2011stanforddogs}, (ii) coarse-grained classification on CIFAR-100~\cite{krizhevsky2009learning}, (iii) object counting on CLEVR~\cite{johnson2017clevr}, and (iv) monocular distance estimation using KITTI~\cite{geiger2013vision}. We compared INFL against the four baselines introduced in the main text and further included a homomorphic-encryption variant (FL-HE) instantiated on a Vision Transformer (ViT) backbone~\cite{dosovitskiy2021an}. Unlike the pretraining protocol used elsewhere in this paper, here we adopt a parameter-efficient finetuning regime based on linear probing (LP)~\cite{alain2017understanding} as a standard PEFT setting~\cite{hu2022lora} to demonstrate that INFL remains effective across diverse training modalities.

\begin{figure}[t]
\renewcommand{\figurename}{Supplementary Fig.}
  \centering
  \includegraphics[width=\textwidth]{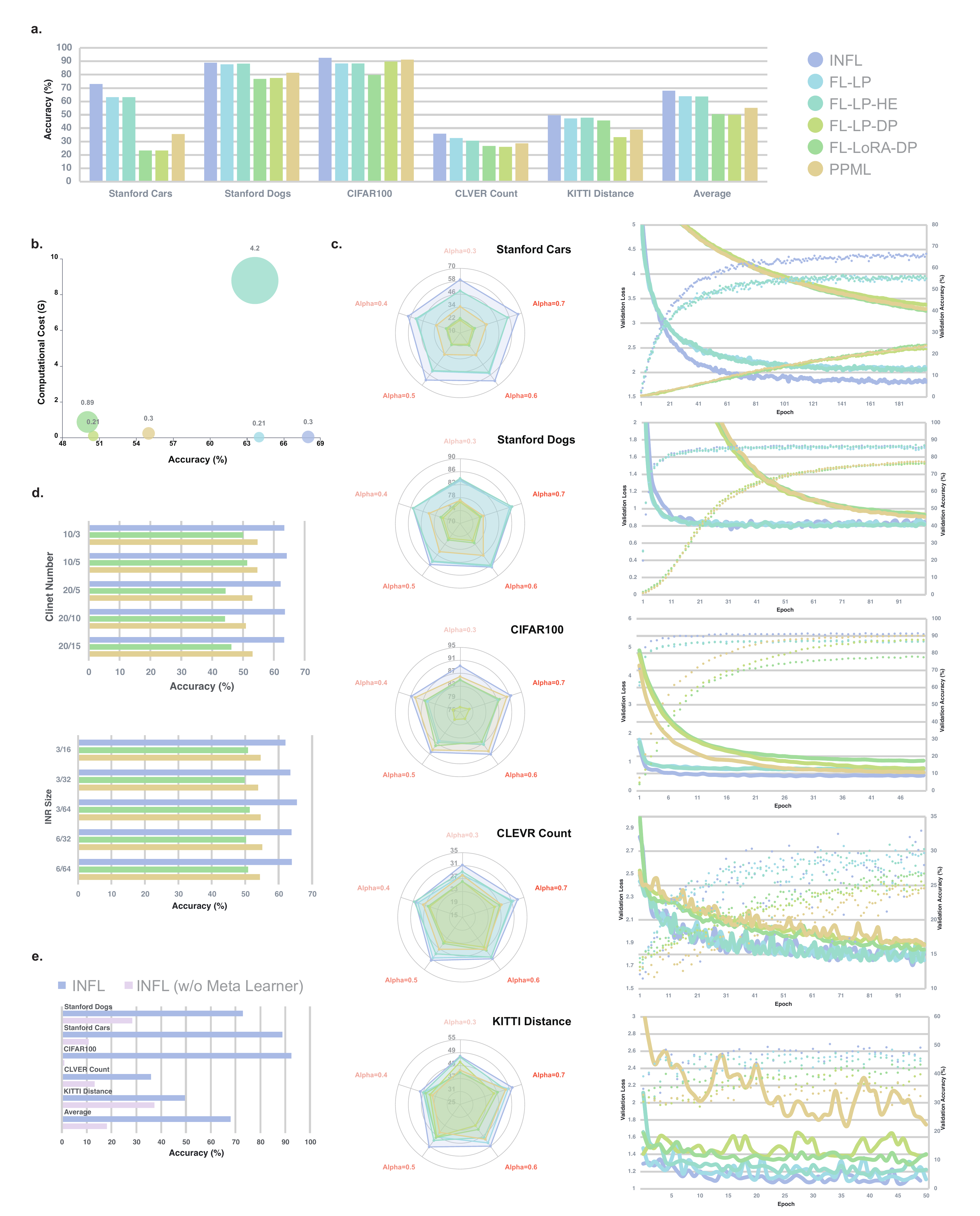}
  \caption{
  \textbf{Sanity check on general classification tasks.}
  \textbf{a} Performance comparison of INFL against four baselines (FL-LP, FL-LP-HE, FL-LP-DP, FL-LoRA-DP, PPML) across five benchmarks (Stanford Cars, Stanford Dogs, CIFAR-100, CLEVR Count, and KITTI Distance), showing accuracy (bar charts).
  \textbf{b} Computational cost versus accuracy trade-off analysis, highlighting INFL's efficiency compared to the high-overhead FL-LP-HE baseline.
  \textbf{c} Robustness evaluation against data heterogeneity under different non-IID levels (Dirichlet concentration $\alpha \in \{0.3, \dots, 0.7\}$) and convergence curves (right column).
  \textbf{d} Ablation studies examining the impact of varying client numbers (top) and INR sizes (bottom) on model performance.
  \textbf{e} Privacy stress test comparing the standard INFL against a variant without the meta-learner (simulating an unauthorized attack), demonstrating the necessity of the key-conditioned design for maintaining utility.
  }
  \label{fig:figure1-supplementary}
\end{figure}

First, INFL attains the highest accuracy and the fastest convergence across all five benchmarks (\textbf{Supplementary Figure~\ref{fig:figure1-supplementary}a–b}), with especially pronounced gains over DP-based baselines on fine-grained recognition (Stanford Cars). These results suggest that the plug-in INR modules capture subtle, class-discriminative cues without the utility losses typically associated with noise-injected training. Notably, INFL also surpasses FL-LP, which does not incur explicit privacy-driven degradation and matches or exceeds FL-LP-HE while avoiding the heavy cryptographic overhead. In particular, whereas FL-LP-HE introduces substantial runtime and compute costs (\textbf{Supplementary Figure~\ref{fig:figure1-supplementary}b}), INFL adds less than 0.1\% incremental overhead relative to FL-LP on our setup, underscoring a favorable accuracy–efficiency trade-off.

We further ablate three factors—data heterogeneity (non-IID level), client count, and INR capacity alongside a privacy stress test (\textbf{Supplementary Figure~\ref{fig:figure1-supplementary}c–d}). For non-IID splits, we employ Dirichlet partitioning with concentration $\alpha \in \{0.3, 0.4, 0.5, 0.6, 0.7\}$ following standardfederated learningpractice~\cite{hsu2019measuring}. Across all $\alpha$ and datasets, INFL consistently outperforms competing methods, indicating robustness to cross-client distribution shift. Varying the number of clients and the INR size on Stanford Cars yields the same conclusion: INFL maintains a clear margin over FL-LoRA~\cite{hu2022lora} and PPML baselines, suggesting that the INR provides a stable and scalable adaptation channel under federation.

Finally, we probe privacy by considering an aggressive attacker that discards the meta-learner and attempts direct inference without the correct coordinate key (\textbf{Supplementary Figure~\ref{fig:figure1-supplementary}c–d}). In practical conditions, unauthorized guessing of the private coordinate key collapses INR modulation, driving accuracy to near-zero. Consistent with this expectation, INFL without the meta-learner fails to deliver usable performance (e.g., CIFAR-100 drops to $\approx 0$ and remains below PPML across tasks). These findings reinforce that INFL’s key-conditioned INR design provides strong protection while preserving utility, delivering state-of-the-art performance on privacy-preserving vision tasks with negligible computational overhead.

\end{appendices}

\clearpage
\bibliography{sn-bibliography}

\end{document}